\documentclass[lettersize,journal]{IEEEtran}
\usepackage{amsmath,amsfonts}
\usepackage{algorithmic}
\usepackage{algorithm}
\usepackage{array}
\usepackage[caption=false,font=footnotesize,labelfont=rm,textfont=rm]{subfig}
\usepackage{textcomp}
\usepackage{stfloats}
\usepackage{url}
\usepackage{verbatim}
\usepackage{graphicx}
\usepackage{cite}
\usepackage{graphicx}
\usepackage{amsmath}
\usepackage{makecell}
\usepackage{mathrsfs}
\usepackage{gensymb}
\usepackage[table]{xcolor}
\usepackage{footnote}
\usepackage{amssymb}
\usepackage{multirow}
\usepackage{diagbox}
\usepackage{subcaption}
\usepackage{stfloats}
\usepackage{pifont}
\usepackage{tablefootnote}
\usepackage{array}
\usepackage{lipsum}
\usepackage{arydshln}

\usepackage{xspace}

\begin{document}

\title{TrackletGait: A Robust Framework for \\ Gait Recognition in the Wild}

\author{Shaoxiong~Zhang,
	Jinkai~Zheng, 
	Shangdong~Zhu, 
	Chenggang~Yan% <-this % stops a space
	
	\thanks{Shaoxiong~Zhang is with the School of Communication Engineering, Hangzhou Dianzi University, Hangzhou, China, and also with Key Laboratory of Micro-nano Sensing and IoT of Wenzhou, Wenzhou Institute of Hangzhou Dianzi University, Wenzhou, China.
	E-mail: zhangsx@hdu.edu.cn}% <-this % stops a space

	\thanks{Jinkai~Zheng is with the School of Communication Engineering, Hangzhou Dianzi University, Hangzhou, China, and with Key Laboratory of Micro-nano Sensing and IoT of Wenzhou, Wenzhou Institute of Hangzhou Dianzi University, Wenzhou, China, and also with Lishui Institute of Hangzhou Dianzi University, Lishui, China.
	E-mail: zhengjinkai3@hdu.edu.cn}% <-this % stops a space

	\thanks{Shangdong~Zhu and Chenggang~Yan are with the School of Communication Engineering, Hangzhou Dianzi University, Hangzhou, China.
	E-mail: \{zhushd, cgyan\}@hdu.edu.cn}% <-this % stops a space

	\thanks{Corresponding author: Jinkai~Zheng, Chenggang Yan.}
	\thanks{The code will be released at https://github.com/zhangsx26/TrackletGait}
	\thanks{Manuscript received xxx xx, 2024; revised xxx xx, 2024.}
	}

\markboth{IEEE xxx, VOL. xx, 20xx}%
{Shell \MakeLowercase{\textit{et al.}}: A Sample Article Using IEEEtran.cls for IEEE Journals}


\maketitle

\begin{abstract}
Gait recognition aims to identify individuals based on their body shape and walking patterns. 
Though much progress has been achieved driven by deep learning, gait recognition in real-world surveillance scenarios remains quite challenging to current methods.
Conventional approaches, which rely on periodic gait cycles and controlled environments, struggle with the non-periodic and occluded silhouette sequences encountered in the wild.
In this paper, we propose a novel framework, \textit{TrackletGait}, designed to address these challenges in the wild.
We propose Random Tracklet Sampling, a generalization of existing sampling methods, which strikes a balance between robustness and representation in capturing diverse walking patterns. 
Next, we introduce Haar Wavelet-based Downsampling to preserve information during spatial downsampling.
Finally, we present a Hardness Exclusion Triplet Loss, designed to exclude low-quality silhouettes by discarding hard triplet samples.
TrackletGait achieves state-of-the-art results, with 77.8\% and 80.4\% rank-1 accuracy on the Gait3D and GREW datasets, respectively, while using only 10.3M backbone parameters.
Extensive experiments are also conducted to further investigate the factors affecting gait recognition in the wild.

\end{abstract}


\section{Introduction}
\label{sec:intro}

\IEEEPARstart{G}{ait} recognition aims to identify pedestrians based on their body shapes and walking patterns. Compared to other biometrics, such as face, fingerprint, and iris, gait can be captured from a distance using off-the-shelf sensors in a covert manner. Therefore, gait recognition holds significant potential for a wide range of applications in video surveillance. For instance, prior research has demonstrated the feasibility of gait recognition in criminal investigations \cite{Muramatsu2013}.

\begin{figure}[t]
	\centering
	\includegraphics[width=1\linewidth]{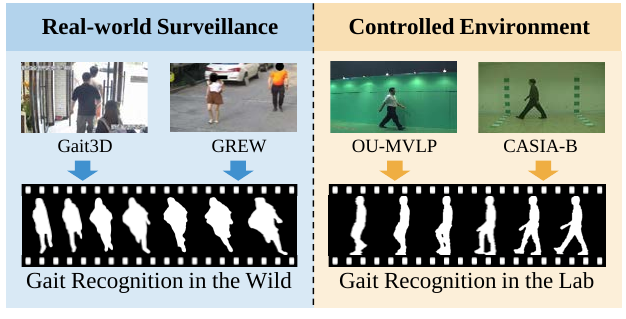}
	\vspace{-0.6cm}
	\caption{Dataset collection scenarios: Gait recognition in the wild vs. gait recognition in the lab.}
	\label{fig:intro}
	\vspace{-0.6cm}
\end{figure}

However, gait recognition has not been widely adopted in video surveillance systems because, when applied to real-world scenarios, its performance tends to decline dramatically \cite{zhu2021gait, zheng2022gait3d}. In fact, gait recognition in such uncontrolled environments remains under-researched. Prior to 2021, research on gait recognition primarily focused on datasets captured in controlled environments, where static cameras are positioned against simple backgrounds \cite{sarkar2005humanid, hofmann2014tum, Yu2006a} or even in front of a green screen \cite{iwama2012isir, Takemura2018}, commonly referred to as gait recognition in the lab. With the release of large gait datasets collected from real-world video surveillance systems, such as GREW \cite{zhu2021gait} and Gait3D \cite{zheng2022gait3d}, new challenges have emerged, referred to as gait recognition in the wild (see Figure~\ref{fig:intro}).

The challenges of gait recognition in the wild can be summarized as follows: conventional gait datasets assume that the subject's walking direction is well-controlled and that walking speed remains unchanged. Consequently, conventional gait silhouette sequences are periodic and align with gait cycles of biomedical. In contrast, in real-world scenarios, pedestrians may walk in arbitrary states, including stopping, turning, and other movements that result in non-periodic silhouette sequences. These differences contribute to the performance degradation of prior deep learning-based gait recognition methods.

In 2023, DeepGaitV2 \cite{fan2023exploring} was proposed based on the OpenGait project \cite{opengait}, which utilizes a deeper network and residual units to enhance performance for gait recognition in the wild. Results indicate that DeepGaitV2 raises rank-1 accuracy by over 10\% on both Gait3D and GREW, providing an encouraging solution for gait recognition in the wild. However, deeper and larger networks do not improve performance indefinitely. According to the experimental results of DeepGaitV2-P3D \cite{fan2023exploring}, replacing a 64-channel network with a 128-channel network results in only a 0.6\% increase in rank-1 accuracy on Gait3D, while the number of parameters in the backbone network quadrupled from 11.1M to 44.4M. This result indicates that we may not rely on increasing the network size to improve performances; instead, we need to explore more effective feature extraction methods to achieve better performance while maintaining the number of network parameters.

To address the challenges while limiting the number of network parameters, we explore gait recognition in the wild from three aspects: temporal sampling, spatial downsampling, and loss function. First, due to the complexity of walking states, gait silhouettes can vary in speed, viewpoint, and walking condition. Consequently, captured silhouettes are not perfectly periodic. Conventional temporal sampling methods often sample several consecutive frames to extract temporal features during training. Compared to the full sequence, the sampled frames are shorter and focus only on a local time period, ignoring other frames during which walking states change. Therefore, conventional temporal sampling methods make it difficult to extract the most discriminative features from a non-periodic gait sequence. We believe that increasing sampling points is necessary to cover a broader range of walking states.

A special temporal sampling method proposed by GaitSet \cite{Chao11} does not consider continuity between frames. GaitSet randomly samples several independent frames from a gait sequence. This approach allows for the sampling of all positions, thereby covering all walking states within the gait sequence. However, the disadvantage of this method is that it does not extract inter-frame information, which may lead to performance degradation. We believe that the minimum sample length must be maintained to capture sufficient temporal information. 

Inspired by these sampling methods, we plan to randomly sample short tracklets from the original gait silhouette sequences during the training phase, rather than independent frames. A short tracklet contains more than three frames to preserve local temporal information. Furthermore, the shorter the length of a sequence, the more tracklets can be sampled {with a fixed total sampling frames}, allowing us to cover a greater variety of walking states. Our proposed temporal sampling method generalizes two previous sampling methods. Therefore, we can balance robustness and representation of gait features to help address the challenges of gait recognition in the wild.

For spatial downsampling, since gait recognition relies on the extraction of fine details from the edges of pedestrian silhouettes, it is crucial to preserve more spatial details. However, conventional downsampling methods, including 2D max pooling, average pooling, and strided convolution, may result in information loss. To address this, we introduce a lossless information transformation method, Haar Wavelet-based Downsampling \cite{xu2023haar}, into our gait recognition framework for spatial downsampling. This method preserves spatial information as much as possible and facilitates the extraction of more discriminative features.

Some silhouettes from real-world surveillance are not discriminative due to severe occlusions or background subtraction errors. These silhouettes may contain incorrect information and are not useful during the model training. To address this problem, we propose Hardness Exclusion Triplet Loss, an extension of the original triplet loss. Our new loss function disregards {a triplet sample} if the {anchor-to-positive distance} exceeds a specified threshold, prompting the model to exclude abnormal samples with extremely poor silhouette quality.

Based on the above observations and solutions, we propose a new framework for gait recognition in the wild, named TrackletGait. {TrackletGait achieves the state-of-the-art performance on two in-the-wild and two hybrid gait datasets, and demonstrates competitive performance on two in-the-lab datasets.}

Our main contributions are summarized as follows. 
\begin{itemize}
	\item We propose a novel temporal sampling approach, Random Tracklet Sampling, to sample short tracklets rather than long sequences or independent frames from a gait sequence during the training stage.

	\item We apply Haar Wavelet-based Downsampling as a method for spatial downsampling, which decreases spatial resolution while preserving full spatial information.
	
	\item We improve the original triplet loss to Hardness Exclusion Triplet Loss, which is designed to discard hard triplets that are not discriminative in the wild.
	
	\item We conduct experiments on {six} public gait datasets, including both in-the-wild and in-the-lab datasets, to demonstrate the effectiveness of our proposed framework. In Additional, we conduct extensive experiments to further investigate the factors affecting gait recognition in the wild.	
\end{itemize}

%--------------------------------------------------------------------------
\section{Related Work}
\label{sec:related}

In this section, we review previous deep learning-based methods for gait recognition. Deep learning-based methods are generally divided into several categories: appearance-based and model-based. Appearance-based methods only use pedestrian binary silhouettes for identification, while model-based methods incorporate the physical structure of the pedestrian's body using human pose estimation \cite{teepe2022towards,xu2021gait,liu2022symmetry,li2022strong,Fu2023GPGait, deng2023human}. In this paper, we only focus on appearance-based methods.

Appearance-based gait recognition methods utilize Convolutional Neural Networks (CNNs) to extract discriminative features from pedestrian silhouettes. Based on how they operate on the temporal dimension, these methods can be further classified into distinct categories corresponding to different stages of research.

\subsection{Template-based Methods}

Template-based methods simplify temporal calculations by condensing the entire gait sequence into a single template image. The Gait Energy Image (GEI) \cite{han2005individual} is the most widely used template. Wu \textit{et al.} \cite{Wu2017} introduced a three-layer CNN to extract features from GEI, establishing a deep learning-based baseline for gait recognition.

\subsection{Set-based Methods}

Gait template images are calculated from binary silhouette images, which may lead to the loss of detailed information before convolutional operations. To address this limitation, GaitSet is proposed  \cite{Chao11}, which treats the gait cycle as a set of unordered silhouettes. GaitSet independently extracts single-frame features using a shallow CNN, followed by temporal pooling to fuse these features. This approach significantly improves cross-view recognition performance on the OU-MVLP dataset \cite{Takemura2018}.

{Set-based gait recognition methods have become prominent due to their simple network architecture, reduced computational requirements, and superior recognition performance. As a result, various improvements have been proposed, including horizontal part weight \cite{wu2020condition}, feature pyramid \cite{hou2020gait}, residual connection \cite{hou2021set}, attention mechanism \cite{li2022gaitslice}, quality weight \cite{hou2022gait}, and spatiotemporal aggregation \cite{chen2022understanding}. Finally, OpenGait Project \cite{opengait} is established as a new baseline of deep learning-based gait recognition framework.}

\subsection{Sequence-based Methods}

Some methods~\cite{Zhang2019b, Fan2020, Lin2020, huang20213d, lin2021gait} consider not only individual frame silhouettes but also the relationships between adjacent frames in a sequence, rather than treating frames as an unordered set. Long Short-Term Memory (LSTM) networks \cite{Feng2017, Zhang2019b} and three-dimensional convolutional neural networks (3D-CNN) \cite{lin2021gait, huang20213d, ma2023dynamic} are common frameworks for temporal feature extraction. The most representative of these is GaitGL \cite{lin2021gait}, which combines 2D and 3D convolutions to extract spatiotemporal gait embeddings, and uses part-based local branches to capture more discriminative features, leading to significant performance improvements. By leveraging both frame-level spatial and temporal information, these methods provide superior gait representations.

\subsection{{Gait Recognition for Complex Scenarios}}

After 2021, {several new gait datasets, collected from more complex scenarios, have been released}, including the GREW \cite{zhu2021gait}, Gait3D \cite{zheng2022gait3d}, CCPG \cite{li2023depth}, and SUSTech1K \cite{shen2023lidargait}. Experimental results on these datasets reveal that the recognition accuracies of previous models, such as GaitSet \cite{Chao11}, GaitGL \cite{lin2021gait}, and GaitBase \cite{opengait}, are insufficient for  {large-scale} real-world scenarios. These in-the-wild datasets introduce new challenges to gait recognition. To address this challenge, DeepGaitV2 \cite{fan2023exploring} was proposed in 2023, significantly improving gait recognition accuracy in real-world scenarios. DeepGaitV2 utilizes 3D and pseudo-3D residual blocks with a deeper network architecture. As a result, it has become the state-of-the-art benchmark for gait recognition in the wild.
 
{Besides DeepGaitV2, several other methods have been proposed to address the aforementioned challenges. Multi-modality-based methods incorporate one or more data modalities beyond single silhouettes to extract additional information from pedestrians. These modalities include 3D human body reconstruction \cite{li2022multi}, point clouds \cite{shen2023lidargait}, human parsing \cite{wang2023gaitparsing, zheng2023parsing}, RGB video \cite{ye2024biggait}, and the fusion of silhouette and skeleton \cite{cui2023multi, zheng2024gaitstr, fan2024skeletongait}. Due to the potential inaccuracies of silhouettes caused by external noises, incorporating additional information helps correct silhouette errors and enhance gait recognition performance in real-world scenarios. However, these methods require access to RGB surveillance, which increases complexity compared to silhouette-based approaches and may raise data privacy concerns.}

{On the other hand, some studies have attempted to address the challenge of gait recognition in complex scenarios by developing new deep learning frameworks, including progressive learning \cite{dou2022gaitmpl}, counterfactual intervention learning \cite{dou2023gaitgci}, sparse-view clustering \cite{wang2023causal}, body structure clustering \cite{Wang2023HSTL}, neural architecture search \cite{guo2022gait, dou2024clash}, causality-based contrastive learning \cite{xiong2025causality}, large vision models \cite{ye2024biggait}, quality assessment \cite{wang2024qagait}, prompt pools \cite{ma2024learning}, and relation descriptors \cite{wang2025free}. However, the recognition accuracies of these models do not significantly outperform DeepGaitV2, nor do they achieve smaller model parameters or reduced computational requirements on real-world datasets. As a result, DeepGaitV2 remains a strong benchmark for gait recognition in real-world scenarios.}

{Therefore, in this work, we choose to employ the conventional silhouette-based framework and explore temporal sampling, spatial downsampling, and loss functions to achieve the state-of-the-art gait recognition performance in real-world scenarios, while also outperforming the DeepGaitV2 model in both model parameters and computational requirements.}

\begin{figure*}[t]
	\centering
	\includegraphics[width=0.9\linewidth]{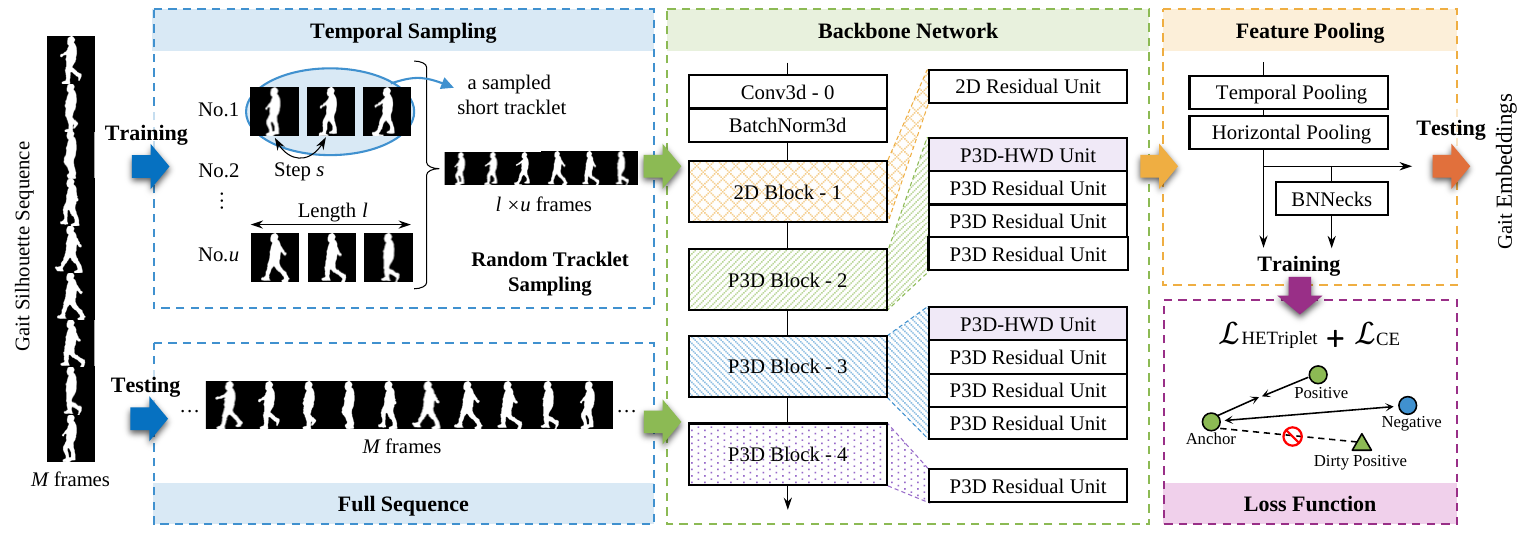}
	\caption{Framework of the proposed TrackletGait.}
	\label{fig:framework}
	\vspace{-0.6cm}
\end{figure*}

\section{Method}
\label{sec:method}
To address gait recognition in the wild, we develop a novel method named \emph{TrackletGait}. The framework of TrackletGait is illustrated in Figure~\ref{fig:framework}. In the following subsections, we will describe each component in detail.

\subsection{Training Temporal Sampling} 
\label{section:sampling}

Given a gait silhouette sequence with $M$ frames $X = \{x_{t}\}$, where $t \in \{1, 2, ..., M\}$, a mini-batch samples $N$ frames $(N\leq M)$ from $X$ for training, denoted as $\mathcal{X}$. Typically, a \textit{Consecutive Sampling} (CS) strategy is adopted:

\begin{equation}
	\mathcal{X}_{\text{CS}} = \{x_{r}, x_{r+1}, ..., x_{r+N-1}\},
\end{equation} 
where $r$ is a random number from $[1, M-N+1]$. In controlled environments, pedestrians tend to walk along a straight line at a uniform speed, making consecutive frames effective for capturing their gait cycle.

In addition to Consecutive Sampling, \textit{Random Frame Sampling} (RFS) {is proposed in GaitSet \cite{Chao11}, based on the consideration that silhouettes, which capture spatial information of pedestrians, tend to remain stable throughout gait cycles, thereby allowing temporal continuity to be disregarded.} Consequently, $N$ independent random frames can be sampled instead of consecutive frames:

\begin{equation}
	\mathcal{X}_{{\text{RFS}}} = \{x_{r_1}, x_{r_2}, ..., x_{r_{N}} \},
\end{equation} 
where {$r_1, r_2, ..., r_{N}$} are {$N$} independent random numbers selected from $[1, M]$. This sampling strategy has also demonstrated improvements on conventional gait benchmarks \cite{Chao11}.

However, in real-world surveillance scenarios, the walking patterns of pedestrians are significantly more complex. {Noise and occlusion can degrade the quality of single-frame silhouettes. Additionally, even within the same silhouette sequence, gait cycles may exhibit non-periodicity. For instance, when a pedestrian walks toward the camera, the size of the silhouettes increases, further contributing to the non-periodicity. Therefore, in these complex scenarios, the two previous sampling strategies face distinct challenges. The CS method struggles to capture the diversity of sequences within a single batch, while the RFS method is heavily influenced by single-frame noise, which results in a decline in recognition performance.}

\begin{figure}[t]
	\centering
	\includegraphics[width=0.9\linewidth]{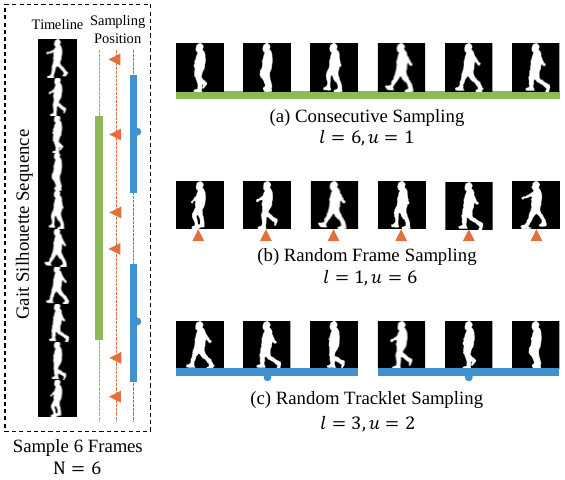}
	\caption{Illustration of three sampling methods.}
	\label{fig:sampling}
	\vspace{-0.5cm}
\end{figure}

To achieve better coverage of variations and a comprehensive representation of between-frame correlations, we propose a generalized and novel sampling method, named \textit{Random Tracklet Sampling} (RTS), as illustrated in Figure~\ref{fig:sampling}. We sample $l$ consecutive frames starting from a randomly selected index within the sequence, repeating this process $u$ times to obtain a total of $N$ frames{, where $l$ is a random variable defined by a given probability distribution}. This method allows us to sample gait data at the tracklet level, {ensuring that sufficient motion information is preserved, while sampling from random positions within the sequence ensures diversity.}
      
Formally, a single tracklet $\mathcal{X}_r$ is a short clip of $X$, containing $l$ frames with step size $s$:		
\begin{equation}
	\mathcal{X}_r = [x_r, x_{r+s}, ..., x_{r+(l-1)s}],
\end{equation}
where $r$ is a random {index}. {Then}, $u$ tracklets are sampled to ensure that the total number of frames reaches $N$, satisfying $ul = N$:	
\begin{equation}
	\mathcal{X}_{\text{RTS-}l} = \mathcal{X}_{r1} \cup \mathcal{X}_{r2} \cup ... \cup \mathcal{X}_{ru}.
\end{equation} 
Finally, $\mathcal{X}_{\text{RTS}}$ is obtained by mixing $\mathcal{X}_{\text{RTS-}l}$ with different values of $l$ according to a probability distribution $P(l)$:
\begin{equation}
	\mathcal{X}_{\text{RTS}} = \mathcal{X}_{\text{RTS-}l}, \quad l \sim P(l).
\end{equation} 
Since $N$ is a fixed integer during the training phase, here $P(l)$ follows a discrete distribution. For example, with $N=32$, the probability mass function of $P(l)$ is given by:
\begin{equation}
	P(l) =
	\begin{cases}
		p_{8}, & \text{if } l = 8 \\
		p_{16},  & \text{if } l = 16 \\
		p_{32}, & \text{if } l = 32 \\
	\end{cases}
\end{equation} 

RTS is a generalization of CS and RFS. By setting {$p_{N}=1, {s=1}$}, it equals to CS. Conversely, with { $p_{1}=1, {s=1}$}, we acquire RFS. Therefore, RTS effectively balances robustness and representation of gait features, helping to address the challenges of gait recognition in the wild.

\subsection{Backbone Network}
\label{sec:backbone}

\begin{figure}[t]
	\centering
	\rotatebox{90}{\includegraphics[width=0.6\linewidth]{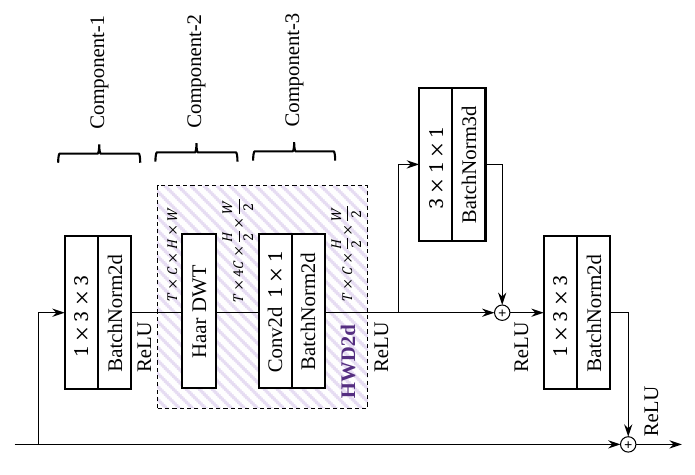}}
	\vspace{-0.3cm}
	\caption{P3D-HWD residual unit.}
	\label{fig:unit}
	\vspace{-0.3cm}
\end{figure}

In previous works on gait recognition in the lab, shallow CNN architectures are typically employed as backbone networks for feature extraction \cite{Wu2017, Chao11}, as a binary gait silhouette is relatively simple compared to chromatic images in other computer vision tasks. However, for gait recognition in the wild, deeper CNN networks have proven to be more effective, especially those using the Residual Network (ResNet) \cite{He2016} architectures, as concluded in DeepGaitV2 \cite{fan2023exploring}. Therefore, we adopt the design of DeepGaitV2  and select a 22-layer ResNet (with unit numbers of $[1,4,4,1]$) as our backbone network. Table~\ref{table_resnet} presents the architecture of our backbone network, where 2D block-1 contains one 2D residual unit, P3D-HWD block-2 and block-3 contain one P3D-HWD residual unit and three P3D residual units, and P3D-HWD block-4 contains one P3D-HWD residual unit. P3D-HWD residual unit is shown in Figure \ref{fig:unit}.

\begin{table}[t]
	\setlength{\abovecaptionskip}{1pt}
	\renewcommand{\arraystretch}{1.3}
	\setlength{\tabcolsep}{10mm}
	\caption{Architecture of the backbone network.}
	\vspace{2mm}
	\setlength{\tabcolsep}{2.5mm}
	\centering
	\begin{tabular}{c|c|c}
		\hline
		\textbf{Layers} & \textbf{Output Size} & \textbf{Block Structure} \\ \hline
		Conv - 0  &  $T\times {\frac{C}{4}}\times64\times44$       & $1\times 1, stride 1$  \\ \hline
		2D Block - 1 & $T\times C\times64\times44$    & $\begin{bmatrix} 3\times3, C \\ 3\times3, C \end{bmatrix}\times1$  \\ \hline
		P3D-HWD Block - 2 & $T\times2C\times32\times22$  & $\begin{bmatrix} 1\times3\times3, 2C \\ 7\times1\times1, 2C \\ 1\times3\times3, 2C \end{bmatrix}\times4$  \\ \hline
		P3D-HWD Block - 3 & $T\times4C\times16\times11$  & $\begin{bmatrix} 1\times3\times3, 4C \\ 7\times1\times1, 4C \\ 1\times3\times3, 4C \end{bmatrix}\times4$  \\ \hline
		P3D-HWD Block - 4 & $T\times8C\times16\times11$  & $\begin{bmatrix} 1\times3\times3, 8C \\ 7\times1\times1, 8C \\ 1\times3\times3, 8C \end{bmatrix}\times1$  \\ \hline
	\end{tabular}
	\label{table_resnet}
	%\vspace{-0.5cm}
\end{table}

\subsubsection{Residual Units}

DeepGaitV2 evaluated the performance of three residual units for gait recognition in both lab and wild environments: 2D, 3D, and P3D residual units. The experiments demonstrate that the P3D residual unit performs competitively with the 3D residual unit. Given that the P3D unit reduces both the number of parameters and FLOPs in the backbone network, DeepGaitV2 recommends selecting the P3D residual unit as a baseline model for gait recognition in the wild to achieve a balance between model efficiency and accuracy. Following this recommendation, we choose the P3D residual unit as the basic component of the backbone network. A kernel size of $7$ and $4$ channel groups are used in our framework, which will be further discussed in Section \ref{sec:discussion}.

\subsubsection{Haar Wavelet-based Downsampling}

DeepGaitV2 employs two strided convolutions for downsampling in P3D Block-2 and Block-3, which may result in information loss. Our proposed TrackletGait utilizes Haar Wavelet-based Downsampling (HWD) \cite{xu2023haar} to perform downsampling without losing spatial information. Specifically, HWD consists of two components: the Haar Discrete Wavelet Transform (Haar DWT) and a standard 2D convolution for channel recovery, as shown in Figure \ref{fig:unit}.

\begin{figure}[t]
	\centering
	\includegraphics[width=0.95\linewidth]{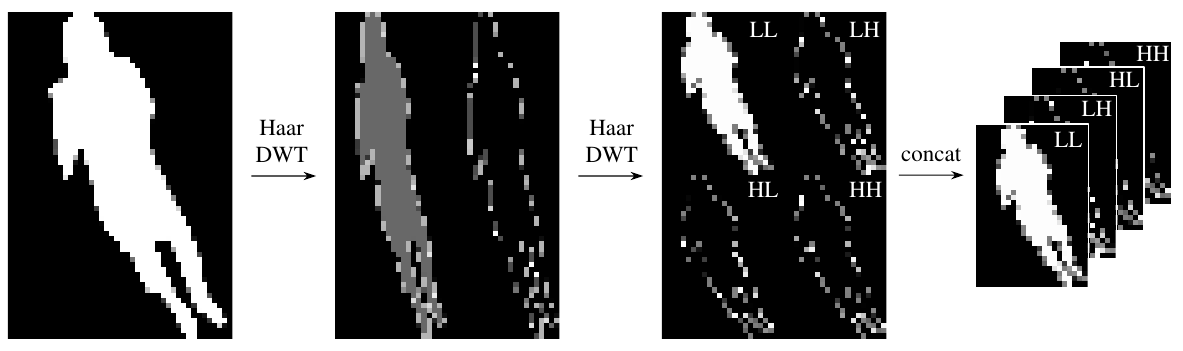}
	\caption{Illustration of the Haar Discrete Wavelet Transform.}
	\label{fig:dwt}
	\vspace{-0.3cm}
\end{figure}

A single-stage Haar transform can be formulated as
\begin{equation}
	\left\{\begin{array}{l}
		\phi_1(x)=\phi_0(2 x)+\phi_0(2 x-1) \\
		\psi_1(x)=\phi_0(2 x)-\phi_0(2 x-1)
	\end{array}\right.
\end{equation}
where $\phi_0$ is given by:
\begin{equation}
	\phi_0(x)=\left\{\begin{array}{cc}
		0, & x<0 \\
		1, & 0 \leqslant x<1 \\
		0, & x \geqslant 1
	\end{array}\right.
\end{equation}
When applying the Haar DWT to a 2D gait silhouette, we perform a single-stage Haar transform separately along both the width and height of the silhouette. After performing these transformations twice, we obtain four components, namely LL, LH, HL, HH, shown in Figure \ref{fig:dwt}. Since $\phi_1(x)$ and $\psi_1(x)$ can be interpreted as low-pass and high-pass filters, these four components—each with half the original size—are described as follows: the approximate low-frequency component (LL), and the high-frequency components, including horizontal details (LH), vertical details (HL), and diagonal details (HH). These four components are concatenated along the channel dimension. As a result, Haar DWT reduces the spatial resolution while increasing the number of channels in the feature maps, thus preserving full information. The size of the feature map is downsampled from $T\times C\times H\times W$ to $T\times 4C\times {\frac{H}{2}}\times {\frac{W}{2}}$. Finally, a standard 2D convolution layer with kernel size of $1\times1$ is applied to recover the original number of channels, resulting in a feature map of size $T\times C\times {\frac{H}{2}}\times {\frac{W}{2}}$.

\subsection{Feature Pooling}

After the backbone network, three common modules are employed, all of which are widely used in conventional gait recognition frameworks.

\subsubsection{Temporal Pooling}
A simple temporal max-pooling layer is applied to compress multiple feature maps into one, as in~\cite{Chao11}. The max-pooling operation helps to extract the most discriminative segments from the full gait sequence for feature extraction and recognition. Notably, since our proposed TrackletGait uses Random Tracklet Sampling, this max-pooling operation also helps fusing information from multiple tracklets.

\subsubsection{Horizontal Pooling}
\label{sec:hpp}
\begin{figure}[t]
	\centering
	\includegraphics[width=\linewidth]{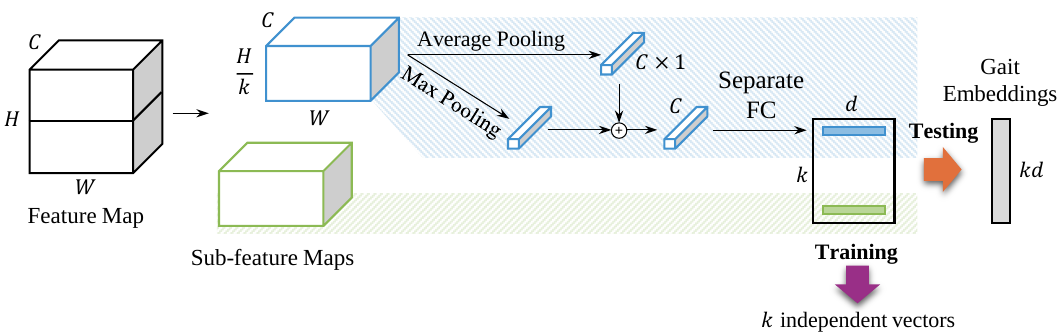}
	\caption{Architecture of the Horizontal Pooling.}
	\label{fig:hpp}
	\vspace{-0.3cm}
\end{figure}

Part-based modules divide the original feature maps into horizontal strips, which have been shown to significantly improve gait recognition performance, as demonstrated by methods such as Horizontal Pyramid Mapping~\cite{Song.2019}, Horizontal Pooling~\cite{Fan2020}, and the Local Feature Extraction Module~\cite{Zhang2019b}. Following these works, we apply Horizontal Pooling in our framework. Similar to DeepGaitV2, our Horizontal Pooling is a simplified version of Horizontal Pyramid Mapping~\cite{fu2019horizontal, Song.2019}, with its architecture shown in Figure~\ref{fig:hpp}.

Horizontal Pooling splits the input feature maps of size $C \times H \times W$ into $k$ horizontal bins, each of size ${C\times \frac{H}{k}}  \times W$. Global Average Pooling (GAP) and Global Max Pooling (GMP) are then applied to downscale each sub-feature map into $C \times 1 \times 1$. These $k$ sub-feature maps of size $C$ are mapped to a new $d$-dimensional feature through separate fully connected networks (FCs). In this work, we set $k=16$ and $d=256$, following DeepGaitV2.

\subsection{Loss Functions}
Batch-all triplet loss \cite{HermansBeyer2017Arxiv} is a commonly used loss function in gait recognition. For each input (anchor), all possible triplets among samples are constructed, including a matching input (positive) and a non-matching input (negative). The Euclidean distance from the anchor to the positive input is minimized, while the distance to the negative input is maximized.

Although batch-all triplet loss works well on gait recognition in the lab, it relies on a crucial assumption: all samples are discriminative, meaning that all pedestrian silhouettes contribute to the recognition. However, in real-world surveillance scenarios, complex environments, noise, and occlusion can render some silhouettes non-discriminative, making them unsuitable for recognition.

\begin{figure}[t]
	\centering
	\includegraphics[width=0.9\linewidth]{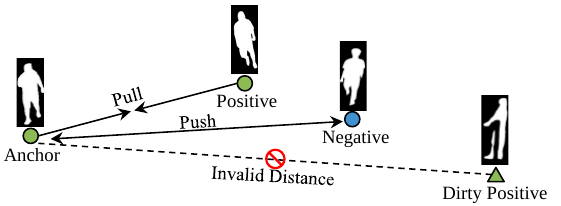}
	\caption{Illustration of the proposed Hardness Exclusion Triplet Loss.}
	\label{fig:loss}
	\vspace{-0.3cm}
\end{figure}

To address this problem, we propose a novel Hardness Exclusion Triplet Loss (HE). The core idea is straightforward: we exclude hard triplets during the training stage (see Figure \ref{fig:loss}). Specifically, we denote a triplet as $(a, p, n)$ where $a$ represents the anchor input, $p$ is the positive input, and $n$ denotes the negative input. The Hardness Exclusion Triplet Loss $\mathcal{L}_{HE}$ is defined as 	
\begin{align}
	\mathcal{L}_{HE} = ReLU(d_{ap}-d_{an}+\xi)\cdot H(d-d_{ap}),
\end{align}
where $d_{ap}$ and $d_{an}$ are the Euclidean distances from the anchor to the positive and negative inputs, respectively. $\xi$ is a margin, $H(\cdot)$ is the Heaviside step function, and $d$ is a distance to distinguish hard triplets in a batch, defined as
\begin{align}
	d = d_{mean} + \alpha(d_{max}-d_{mean}).
	\label{equ:alpha}
\end{align}
where $d_{mean}$ is the mean value of all distances from the anchor to the positive input in a training batch, and $d_{max}$ is the maximum value. $\alpha$ is used to adjust the distance threshold for hard triplets.

In addition, we employ the BNNecks module and use the cross entropy loss $\mathcal{L}_{CE}$, with a default weight of $1$, following OpenGait project~\cite{opengait}.

%---------------------------------------------------------------------------------------- 

\section{Main Experiments}
\label{sec:exp}
In this section, we describe the datasets used in our experiments and present the final results, demonstrating that our proposed method outperforms prior state-of-the-art gait recognition methods.

\begin{table}[t]
	\setlength{\abovecaptionskip}{1pt}
	\renewcommand{\arraystretch}{1.3}
	\caption{{Information on gait datasets utilized in this study. \#ID and \#Seq represent the number of subjects and the number of gait sequences, respectively.}}
	\vspace{2mm}
	\setlength{\tabcolsep}{0.89mm}
	\centering
	\scalebox{0.9}
	{
		\begin{tabular}{c|c|c|c|c|c}
			\hline
			\multirow{2}{*}{\textbf{Dataset}} & \multicolumn{2}{c|}{\textbf{Training Set}} & \multicolumn{2}{c|}{\textbf{Testing Set}} & \multirow{2}{*}{\textbf{Scenario}} \\
			\cline{2-5}
			
			 & \textbf{\#ID} & \textbf{\#Seq.} & \textbf{\#ID} & \textbf{\#Seq.} &  \\ \hline
			Gait3D \cite{zheng2022gait3d} & 3,000 & 18,940 & 1,000 & 6,369 & Real-world Surveillance\\ \hline
			GREW \cite{zhu2021gait,guo2022gait} & 20,000 & 102,887 & 6,000 & 24,000 & Real-world Surveillance\\ \hline
			OU-MVLP \cite{Takemura2018} & 5,153 & 144,284 & 5,154 & 144,412 & Constrained Cross-view\\ \hline
			CASIA-B \cite{Yu2006a} & 74 & 8,140 & 50 & 5,500 & Constrained Cross-view \\ \hline
			{CCPG \cite{li2023depth}} & 100 & 8,187 & 100 & 8,095 & Constrained Clothes-changing \\ \hline
			{SUSTech1K \cite{shen2023lidargait}} & 200 & 5,988 & 850 & 19,228 & Constrained Multi-sensors \\ \hline
			
		\end{tabular}
	}
	\label{tab:datasets}
	\vspace{-0.3cm}
\end{table}

\begin{table*}[ht]
	\setlength{\abovecaptionskip}{1pt}
	\renewcommand{\arraystretch}{1.3}
	\caption{Implementation details. The MultiStepLR scheduler decays the learning rate by a factor of $\gamma$ once the number of training iterations reaches any of the predefined milestones.}
	\vspace{2mm}
	\setlength{\tabcolsep}{5.8mm}
	\centering
	\scalebox{0.9}
	{
	\begin{tabular}{c|c!{\vrule width 1.2pt}c|c|ccc}
		\hline
		\multirow{2}{*}{\textbf{Dataset}} &
		\multirow{2}{*}{\textbf{\#ID}} & \multirow{2}{*}{\renewcommand{\arraystretch}{1.2}
			\begin{tabular}
				[c]{@{}c@{}}\textbf{Batch Size}\\ $\boldsymbol{(p\times k)}$
		\end{tabular}} & 
		\multirow{2}{*}{\textbf{Optimizer}} &
		\multicolumn{3}{c}{\textbf{Training Scheduler}}
		
		\\ 
		\cline{5-7}
		
		& & & & \multicolumn{1}{c|}{\textbf{Type}}  & \multicolumn{1}{c|}{\textbf{Milestones}}   & \textbf{Iterations} \\ \hline
		
		Gait3D \cite{zheng2022gait3d} & 4,000 &$(32\times4)$ & \multirow{6}{*}{\begin{tabular}[c]{@{}c@{}}SGD\\ $Learning Rate=0.1$\\ $Weight Decay=0.0005$\\$Momentum=0.9$\end{tabular}} & \multicolumn{1}{c|}{\multirow{6}{*}{\begin{tabular}[c]{@{}c@{}}MultiStepLR\\ $\gamma=0.1$\end{tabular}}} & \multicolumn{1}{c|}{$40k, 80k, 100k$} & $120k$\\ 
		\cline{1-3} \cline{6-7} 
		
		GREW \cite{zhu2021gait} & 26,000 & $(32\times4)$ & & \multicolumn{1}{c|}{}  & \multicolumn{1}{c|}{$80k, 120k, 150k$} & $180k$ \\
		\cline{1-3} \cline{6-7} 
		
		OU-MVLP \cite{Takemura2018} & 10,307 & $(32\times8)$ & & \multicolumn{1}{c|}{}  & \multicolumn{1}{c|}{$60k, 80k, 100k$}  & $120k$ \\ 
		\cline{1-3} \cline{6-7} 
		
		CASIA-B \cite{Yu2006a} & 124 & $(8\times16)$ & & \multicolumn{1}{c|}{}  & \multicolumn{1}{c|}{$20k, 40k, 50k$}   & $60k$  \\
		\cline{1-3} \cline{6-7} 
		
		{CCPG \cite{li2023depth}} & 200 & $(8\times16)$ & & \multicolumn{1}{c|}{}  & \multicolumn{1}{c|}{$20k, 40k, 50k$}   & $60k$ \\
		\cline{1-3} \cline{6-7} 
		
		{SUSTech1K \cite{shen2023lidargait}} & 1,050 & $(8\times8)$ & & \multicolumn{1}{c|}{}  & \multicolumn{1}{c|}{$20k, 30k, 40k$}   & $50k$ \\
		\hline
		
	\end{tabular}
	}
	\label{table:parameters}
	%\vspace{-0.3cm}
\end{table*}

\subsection{Datasets}
\label{section:dataset}

{We evaluate the performance of the proposed TrackletGait on six public gait datasets, presented in Table \ref{tab:datasets}. Gait3D and GREW are classified as in-the-wild datasets, reflecting real-world surveillance scenarios. In contrast, OU-MVLP and CASIA-B are in-the-lab datasets collected under fully controlled conditions. CCPG and SUSTech1K are hybrid datasets that include outdoor scenes but do not represent real-world surveillance scenarios.}
\subsection{Implementation Details}
We adopt the implementations from the OpenGait project \cite{opengait}, using the default auto-mixed precision mode. {Experiments with $64$-channel are conducted on four NVIDIA GeForce RTX 3090 GPUs, with PyTorch version 1.10.0. Only silhouette images are utilized in this study}, and all input silhouettes are resized to $64 \times 44$. {In Random Tracklet Sampling, we set $N=32, p_{8}=0.2, p_{16}=0.3, p_{32}=0.5$ to achieve a balanced parameter configuration across different scenarios, thereby ensuring consistency across all datasets. In HE Triplet Loss, we set $\xi=0.2, \alpha=2/3$.} Due to variations in data size and scenarios across the datasets, we adjust the other parameter settings for each dataset accordingly, presented in Table \ref{table:parameters}. Additionally, {For the smaller datasets CASIA-B, CCPG, and SUSTech1K, we reduce the number of residual units from $[1,4,4,1]$ to $[0,1,1,0]$ to prevent overfitting.}

\begin{table*}[t]
\setlength{\abovecaptionskip}{1pt}
\renewcommand{\arraystretch}{1.3}
\caption{Comparison on Gait3D \cite{zheng2022gait3d}, GREW \cite{zhu2021gait}, OU-MVLP \cite{Takemura2018}, CASIA-B \cite{Yu2006a}, {CCPG \cite{li2023depth}, and SUSTech1K \cite{shen2023lidargait}} datasets. The identical-view cases are excluded on CASIA-B and OU-MVLP. {The results on CASIA-B and SUSTech1K are the average of all testing subsets.} {R$n$: Rank-$n$ accuracy (\%).}}
\vspace{2mm}
\setlength{\tabcolsep}{1.2mm}
\scalebox{0.90}
{
	\begin{tabular}{c|c|c|c!{\vrule width 1.2pt}c|c|c!{\vrule width 1.2pt}c|c|c!{\vrule width 1.2pt}c!{\vrule width 1.2pt}c!{\vrule width 1.2pt}c|c|c!{\vrule width 1.2pt}c|c}
		\hline
		
		\multirow{2}{*}{\textbf{Method}} & \multirow{2}{*}{\textbf{Source}}& \multirow{2}{*}{{\renewcommand{\arraystretch}{1.2}\begin{tabular}[c]{@{}c@{}}\textbf{Params}$^\dagger$\\ \textbf{(M)}\end{tabular}}} & \multirow{2}{*}{{\renewcommand{\arraystretch}{1.2}\begin{tabular}[c]{@{}c@{}}\textbf{{MACs}$^\ddagger$}\\ {\textbf{(G)}}\end{tabular}}} & \multicolumn{3}{c!{\vrule width 1.2pt}}{\textbf{Gait3D}} & \multicolumn{3}{c!{\vrule width 1.2pt}}{\textbf{GREW}} & \textbf{OU-MVLP} & \textbf{CASIA-B} & \multicolumn{3}{c!{\vrule width 1.2pt}}{\textbf{{CCPG-G}}} & \multicolumn{2}{c}{\textbf{{SUSTech1K}}}\\ 
		\cline{5-17}
		&&&&\textbf{R1}&\textbf{R5}&\textbf{mAP}&\textbf{R1}&\textbf{R5}&\textbf{R10}&\textbf{R1}&\textbf{R1}&\textbf{CL-R1}&\textbf{UP-R1}&\textbf{DN-R1}&\textbf{R1}&\textbf{R5}\\
		
		\hline
		GaitSet~\cite{Chao11}&AAAI-19&2.3&13.7&36.7&58.3&30.0&46.3&63.6&70.3&87.1&84.2&77.7&83.5&83.2&65.0&84.8\\
		GaitPart~\cite{Fan2020}&CVPR-20&1.5&-&28.2&47.6&21.6&44.0&60.7&67.3&88.7&88.8&77.8&84.5&83.3&59.2&80.8\\
		GaitGL~\cite{lin2021gait}&ICCV-21&2.5&-&29.7&48.5&22.3&47.3&63.6&69.3&89.7&91.8&69.1&75.5&77.6&63.1&82.8\\
		
		SMPLGait~\cite{zheng2022gait3d}&CVPR-22&-&-&46.3&64.5&37.2&-&-&-&-&-&-&-&-&-&-\\
		{MetaGait~\cite{dou2022metagait}}&ECCV-22&-&-&-&-&-&-&-&-&91.9&93.4&-&-&-&-&-\\
		{GaitMPL~\cite{dou2022gaitmpl}}&TIP-22&-&-&-&-&-&-&-&-&90.6&93.3&-&-&-&-&-\\
		{GaitStrip~\cite{wang2022gaitstrip}}&ACCV-22&-&-&-&-&-&-&-&-&*&93.0&-&-&-&-&-\\
		
		\hdashline[1pt/1pt]
		GaitBase~\cite{opengait}&CVPR-23&4.9&-&64.6&-&-&60.1&-&-&90.8&89.7&-&-&-&76.1&89.4\\
		DANet~\cite{ma2023dynamic}&CVPR-23&-&-&48.0&69.7&-&-&-&-&90.7&-&-&-&-&-&-\\
		GaitGCI~\cite{dou2023gaitgci}&CVPR-23&-&-&50.3&68.5&24.3&68.5&80.8&84.9&92.1&94.5&-&-&-&-&-\\
		{AUG-OGBase~\cite{li2023depth}}&CVPR-23&-&-&-&-&-&-&-&-&-&-&84.7&88.4&89.4&-&-\\
		DyGait~\cite{Wang2023DyGait}&ICCV-23&-&-&66.3&80.8&56.4&71.4&83.2&86.8&*&94.1&-&-&-&-&-\\
		HSTL~\cite{Wang2023HSTL}&ICCV-23&4.1&-&61.3&76.3&55.5&62.7&76.6&81.3&\textbf{92.4}&94.3&-&-&-&-&-\\
		GaitAMR~\cite{chen2023gaitamr}&IS-23&-&-&-&-&-&-&-&-&88.3&92.5&-&-&-&-&-\\

		DeepGaitV2-3D-64~\cite{fan2023exploring}&arXiv-23&27.5&217.3&72.8&86.2&63.9&79.4&88.9&91.4&92.0&89.6&-&-&-&-&-\\
		DeepGaitV2-P3D-64~\cite{fan2023exploring}&arXiv-23&11.1&91.2&74.4&88.0&65.8&77.7&87.9&90.6&91.9&89.6&91.1&95.3&92.9&82.4&92.6\\
		SwinGait-3D~\cite{fan2023exploring}&arXiv-23&13.1&-&75.0&86.7&67.2&79.3&88.9&91.8&-&-&-&-&-&79.7&91.8\\
		
		\hdashline[1pt/1pt]
		GaitDAN~\cite{huang2024gaitdan}&TCSVT-24&-&-&-&-&-&-&-&-&90.2&93.0&-&-&-&-&-\\
		AttenGait{(sil.)}~\cite{castro2024attengait}&PR-24&-&-&-&-&-&50.2&67.3&73.4&-&85.6
		&-&-&-&-&-\\
		HybridGait~\cite{dong2024hybridgait}&AAAI-24&-&-&53.3&72.0&43.3&-&-&-&-&-&-&-&-&-&-\\
		QAGait~\cite{wang2024qagait}&AAAI-24&-&-&67.0&81.5&56.5&59.1&74.0&79.2&-&90.2&-&-&-&-&-\\
		VPNet~\cite{ma2024learning}&CVPR-24&-&-&75.4&87.1&-&80.0&89.4&-&\textbf{92.4}&\textbf{94.9}&-&-&-&-&-\\
		{CLTD~\cite{xiong2025causality}}&ECCV-24&-&-&69.7&85.2&-&78.0&87.8&-&92.3&94.8&-&-&-&-&-\\	{GaitBase+RD~\cite{xiong2025causality}}&ECCV-24&-&-&70.1&-&61.9&65.5&78.7&83.3&91.3&90.0&92.3&95.5&95.3&-&-\\
		{CLASH~\cite{dou2024clash}}&TIP-24&-&-&52.4&69.2&40.2&67.0&78.9&83.0&91.9&93.9&-&-&-&-&-\\
		\rowcolor[gray]{0.9}
		\textbf{{TrackletGait-64}}&Ours&10.3&88.0&\textbf{77.8}&\textbf{89.0}&\textbf{70.2}&\textbf{80.4}&\textbf{89.8}&\textbf{92.2}&91.9&94.1&\textbf{92.5}&\textbf{96.7}&\textbf{96.0}&\textbf{86.2}&\textbf{94.0}\\

		\hline
		DeepGaitV2-P3D-128~\cite{fan2023exploring}&arXiv-23&44.4&364.0&75.0&-&-&81.0&-&-&-&-&-&-&-&-&-\\
		DeepGaitV2-3D-128~\cite{fan2023exploring}&arXiv-23&109.8&868.7&75.8&-&-&81.6&-&-&-&-&-&-&-&-&-\\
		
		\rowcolor[gray]{0.9}
		\textbf{TrackletGait-128}&Ours&41.2&349.6&\textbf{79.8}&\textbf{90.0}&\textbf{71.8}&\textbf{83.2}&\textbf{91.4}&\textbf{93.5}&-&-&-&-&-&-&-\\
		\hline
	\end{tabular}	
}
\vspace{0.2em}

%\noindent %
\parbox{\linewidth}{%
	\scriptsize
	$*$ Accuracy without invalid probe sequences is presented, which is a different protocol with ours.\\
	{$\dagger$ Only the parameters within the backbone network are taken into consideration.\\
		$\ddagger$ Multiply-Accumulate Operations, calculated using PyTorch-OpCounter with an input size of $1 \times 32 \times 64 \times 44$.}
	}

\label{tab:main_result}
\vspace{-0.5cm}
\end{table*}

\subsection{Main Results}
\label{sec:exp_mainresults}
We compare our TrackletGait method with previous approaches on several gait datasets. The main results are summarized in Table~\ref{tab:main_result}. {It is important to note that, as only silhouette images are used without incorporating pose, RGB, or other modalities, our results are compared with methods that utilize the same data modality to ensure fair comparisons.}

On {Gait3D}, TrackletGait achieves a rank-1 accuracy of 77.8\%, outperforming previous state-of-the-art methods, including DeepGaitV2-3D-128 (+2.0\%) and VPNet-L (+2.4\%). It is noteworthy that the number of backbone parameters in both DeepGaitV2-3D-128 and VPNet-L is significantly higher than that of our method. 
Our TrackletGait-64 ($C=64$) employs only 10.3M parameters, which is ten percent of the parameters used in DeepGaitV2-3D-128. 
As for VPNet-L, since the original paper did not provide the exact number of backbone parameters, we estimate it to be over 60M, as they utilize stacked 3D residual units with numbers of $[3,4,6,3]$. When comparing accuracy with similar backbone networks, DeepGaitV2-P3D-64, we achieve an improvement of +3.4\% in rank-1 accuracy. On {GREW}, with $C=64$ groups, we achieve a rank-1 accuracy of 80.4\%, outperforming DeepGaitV2-P3D-64 (+2.7\%) and DeepGaitV2-3D-64 (+1.0\%). Overall, these experiments demonstrate the simplicity and efficiency of our TrackletGait method for gait recognition in the wild.

{On the in-the-lab datasets, TrackletGait achieves an average rank-1 accuracy of 91.9\% on OU-MVLP and 94.1\% on CASIA-B, which are only 0.5\% and 0.8\% lower than the current state-of-the-art methods. Despite this marginal difference, TrackletGait remains competitive for gait recognition in the lab, especially considering that it is specifically designed for gait recognition in the wild.}
{On two hybrid datasets, CCPG and SUSTech1K, TrackletGait achieves a state-of-the-art performance. This result highlights the generalization capability of TrackletGait, which ensures robust recognition across diverse outdoor scenarios, including variations such as clothing changes, object carrying, and other challenging conditions.}

{Finally, to compare the 128-channel DeepGaitV2-P3D and -3D models, we conducted additional experiments on Gait3D and GREW. In these experiments, the number of channels in TrackletGait is increased to match that of DeepGaitV2-128. The results, shown in the last three rows of Table~\ref{tab:main_result}, demonstrate that TrackletGait significantly outperforms both DeepGaitV2-P3D and -3D, even with lower parameter and computational complexity. These results validate the effectiveness of TrackletGait in real-world surveillance scenarios.}

\subsection{{Extensive Results}}
\label{sec:ExtensiveResults}

{
We present the extensive results in Table~\ref{tab:temporalsampling} to evaluate the performance of TrackletGait under varying sampling lengths, demonstrating the model's best performance potential. In contrast, Table~\ref{tab:main_result} shows the results obtained by fusing these three sampling lengths. Thus, the performances presented in Table~\ref{tab:temporalsampling} may surpass those reported in the main results shown in Table~\ref{tab:main_result}. For gait datasets collected in the wild, the best performance is achieved with RTS-8 on Gait3D, while RTS-32 yields optimal results on GREW. When extending the experiments to in-the-lab and hybrid datasets, RTS-16 demonstrates superior performance. A detailed explanation of these results will be discussed in Section~\ref{sec:SamplingStrategies}.
}

\begin{table}[t]
	\setlength{\abovecaptionskip}{1pt}
	\renewcommand{\arraystretch}{1.3}
	\caption{Extensive experiments considering temporal sampling strategy.}
	\setlength{\tabcolsep}{0.75mm}
	\centering
	
	\begin{tabular}{c|c|c|c|c|c|c}
		\hline
		\multirow{2}{*}{\renewcommand{\arraystretch}{1.2}\begin{tabular}[c]{@{}c@{}}\textbf{Sampling}\\ \textbf{Strategy}\end{tabular}} & \multicolumn{6}{c}{\textbf{Dataset, Average Rank-1 Accuracy (\%)}} \\ 
		\cline{2-7} 	
		& {\textbf{Gait3D}} & {\textbf{GREW}} & {\textbf{OU-MVLP}} & {\textbf{CASIA-B}} & {\textbf{CCPG}} & \textbf{SUSTech1K} \\ 
		\hline	
		RTS-$32$ & 74.8 & \textbf{80.5} & 92.0 & 93.2 & 93.2 & 85.5\\ 	
		RTS-$16$ & 77.4 & 79.9 & \textbf{92.1} & \textbf{94.4} & \textbf{94.7} & \textbf{86.3}\\ 
		RTS-$8$  & \textbf{78.2} & 77.6 & 90.6 & 93.2 & 94.6 & 85.9\\ 
		\hline
	\end{tabular}
	
	\label{tab:temporalsampling}
%	\vspace{-0.3cm}
\end{table}

\subsection{Ablation Results}
\label{sec:ablation}

In Table~\ref{tab:ablation}, we present the ablation results on the Gait3D dataset to demonstrate the effectiveness of the proposed modules in TrackletGait. {We start ablation experiments from a baseline model, which is already stronger than that of DeepGaitV2-P3D-64.} Specifically, we enhanced the temporal branch by replacing the convolutional layer with a $3\times1\times1$ kernel with a larger $7\times1\times1$ kernel, which will be discussed in Section \ref{sec:TemporalBranch}. We do not claim these improvements as the primary contributions of this paper, but rather as tricks to improve performance. Then we observe further performance gains when incorporating the key components of our full model. These findings indicate that while our baseline model benefits from extended training, the proposed modules further improve accuracy.

\begin{table}[t]
	\setlength{\abovecaptionskip}{1pt}
	\renewcommand{\arraystretch}{1.3}
	\caption{{Ablation experiments on {Gait3D} dataset.}}
	\setlength{\tabcolsep}{3.5mm}
	\centering
	\begin{tabular}{c|ccc|c}
		\hline
		\multirow{2}{*}{\renewcommand{\arraystretch}{1.2}\begin{tabular}[c]{@{}c@{}}\textbf{Exp.}\\ \textbf{Index}\end{tabular}} & \multicolumn{3}{c|}{\textbf{Key Component}}  & \multirow{2}{*}{\renewcommand{\arraystretch}{1.2}\begin{tabular}[c]{@{}c@{}}\textbf{Rank-1}\\ \textbf{Accuracy (\%)}\end{tabular}}             \\
		
		\cline{2-4} 
		
		& \textbf{RTS} & \textbf{HWD} & \textbf{HE Triplet} &  \\
		
		\hline
		\ding{172} & $\times$     & $\times $    & $\times$      & 75.9 \\ 
		\ding{173} & $\checkmark$ & $\times$     & $\times$      & 77.0 \\ 
		\ding{174} & $\checkmark$ & $\checkmark$ & $\times$      & 77.2 \\ 
		\ding{175} & $\checkmark$ & $\checkmark$ & $\checkmark$  & \textbf{77.8} \\ 	
		\hline
		
	\end{tabular}
	\label{tab:ablation}
%	\vspace{-0.3cm}
\end{table}

%-------------------------------------------------------------------
\section{Discussion}
\label{sec:discussion}
In this section, we conduct a series of experiments on Gait3D to further investigate the factors affecting gait recognition in the wild and present our insights. We choose DeepGaitV2-P3D-64~\cite{fan2023exploring} as the primary comparison for our proposed TrackletGait, as DeepGaitV2-P3D is a strong and well-balanced state-of-the-art method for gait recognition in the wild. {Since RTS-$8$ achieves the best performance (78.2\%) on Gait3D (see Table~\ref{tab:temporalsampling}), the experiments in this subsection are conducted based on RTS-$8$.}

\subsection{Discussion on Temporal Sampling Strategies}
\label{sec:SamplingStrategies}

{In Section \ref{sec:ExtensiveResults}, we present the results for different sampling lengths. To better understand this observation, we consider the scene variations across different datasets. Table \ref{tab:aifv} provides information on the scenarios, the number of subjects, and the estimation of half gait cycles. We can observe that a notable difference between the wild dataset and constrained datasets (Figure \ref{fig:intro} illustrates their environments) is the gait cycle estimation results, where wild datasets exhibit higher variance.} 

\begin{table}[t]
	\setlength{\abovecaptionskip}{1pt}
	\renewcommand{\arraystretch}{1.3}
	\caption{{Comparison Across Gait Datasets. W: Wild; C: Constrained; I: Indoors; O: Outdoors. Var: Variance.}}
	\setlength{\tabcolsep}{0.5mm}
	\centering
	
	\begin{tabular}{c|c|c|c|c|c|c|c}
		\hline
		\multicolumn{2}{c|}{\multirow{2}{*}{\renewcommand{\arraystretch}{1.2}\begin{tabular}[c]{@{}c@{}}\textbf{Comp.}\\ \textbf{Item}\end{tabular}}} & \multicolumn{6}{c}{\textbf{Dataset}} \\ 
		\cline{3-8} 	
		\multicolumn{1}{c}{}&& {\textbf{Gait3D}} & {\textbf{GREW}} & {\textbf{OU-MVLP}} & {\textbf{CASIA-B}} & {\textbf{CCPG}} & \textbf{SUSTech1K} \\ 
		\hline
		\multicolumn{2}{c|}{\textbf{Scenario}} & W, I & W, O & C, I & C, I & C, I+O & C, O \\ 	
		\hline
		
		\multicolumn{2}{c|}{\textbf{\#ID}} & 4,000 & 26,000 & 10,307 & 124   & 200   & 1,050  \\ 	
		\hline
		
		\multirow{2}{*}{\renewcommand{\arraystretch}{1.2}\begin{tabular}[c]{@{}c@{}}\textbf{Half}\\ \textbf{Cycle}\end{tabular}}  & Mean & 15.8  & 15.2   & 12.6   & 12.7  & 15.0  & 12.6   \\ 
		\cline{2-8} 	
		& Var. & 18.8  & 23.6   & 4.4   & 5.3  & 9.4  & 5.1 \\ 
		\hline
		\multicolumn{2}{c|}{\textbf{AIFV}}       & 0.217 & 0.187  & 0.055  & 0.077 & 0.112 & 0.098  \\ 
		\hline
	\end{tabular}
	
	\label{tab:aifv}
%	\vspace{-0.3cm}
\end{table}

\begin{figure*}[htbp]
	\centering
	\begin{minipage}[b]{0.29\textwidth}
		\centering
		\subfloat[Gait3D]{\includegraphics[width=0.99\textwidth, trim=8cm 2.5cm 5.5cm 2.5cm, clip]{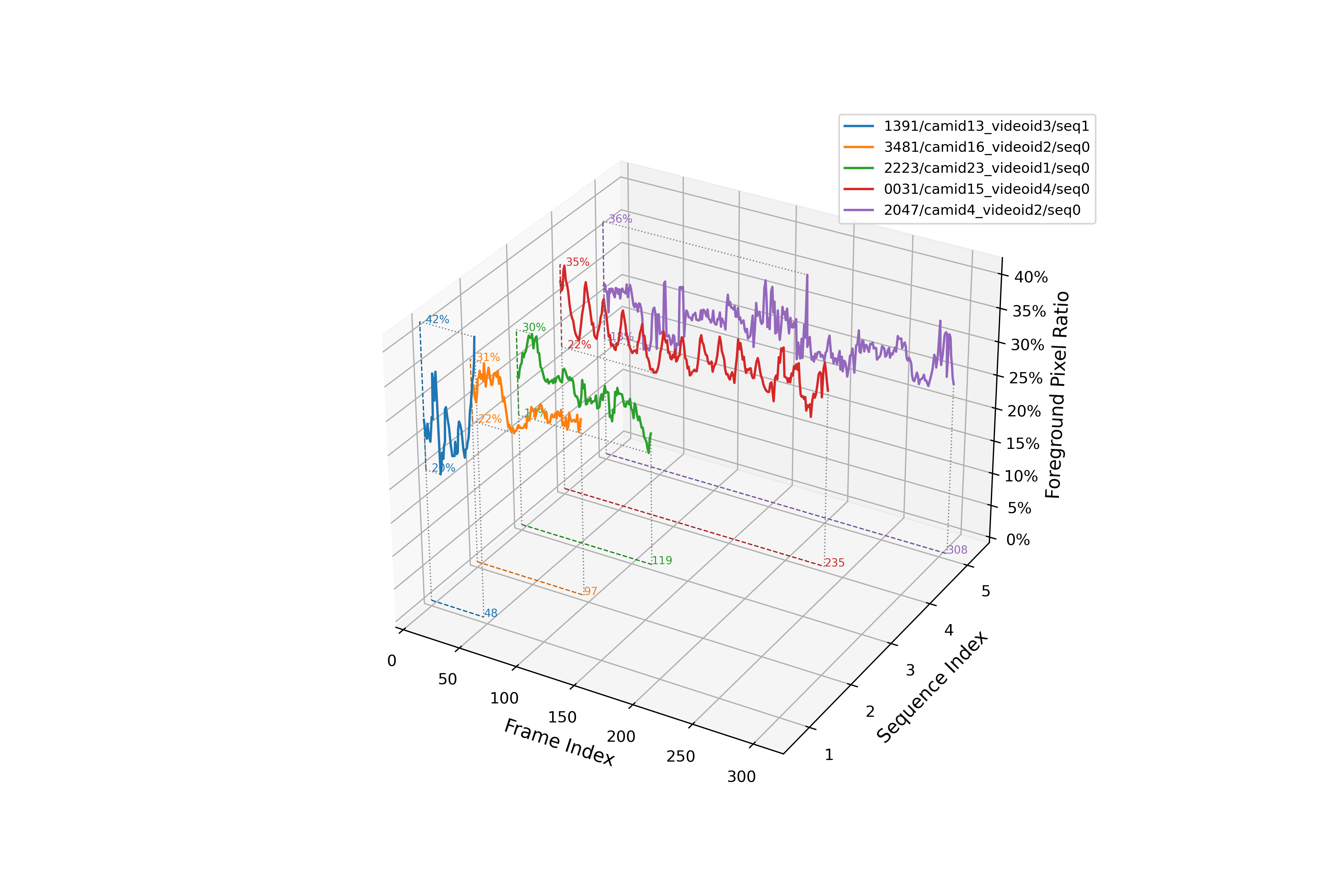}}
	\end{minipage}
	\hfill
	\begin{minipage}[b]{0.29\textwidth}
		\centering
		\subfloat[GREW]{\includegraphics[width=0.99\textwidth, trim=8cm 2.5cm 5.5cm 2.5cm, clip]{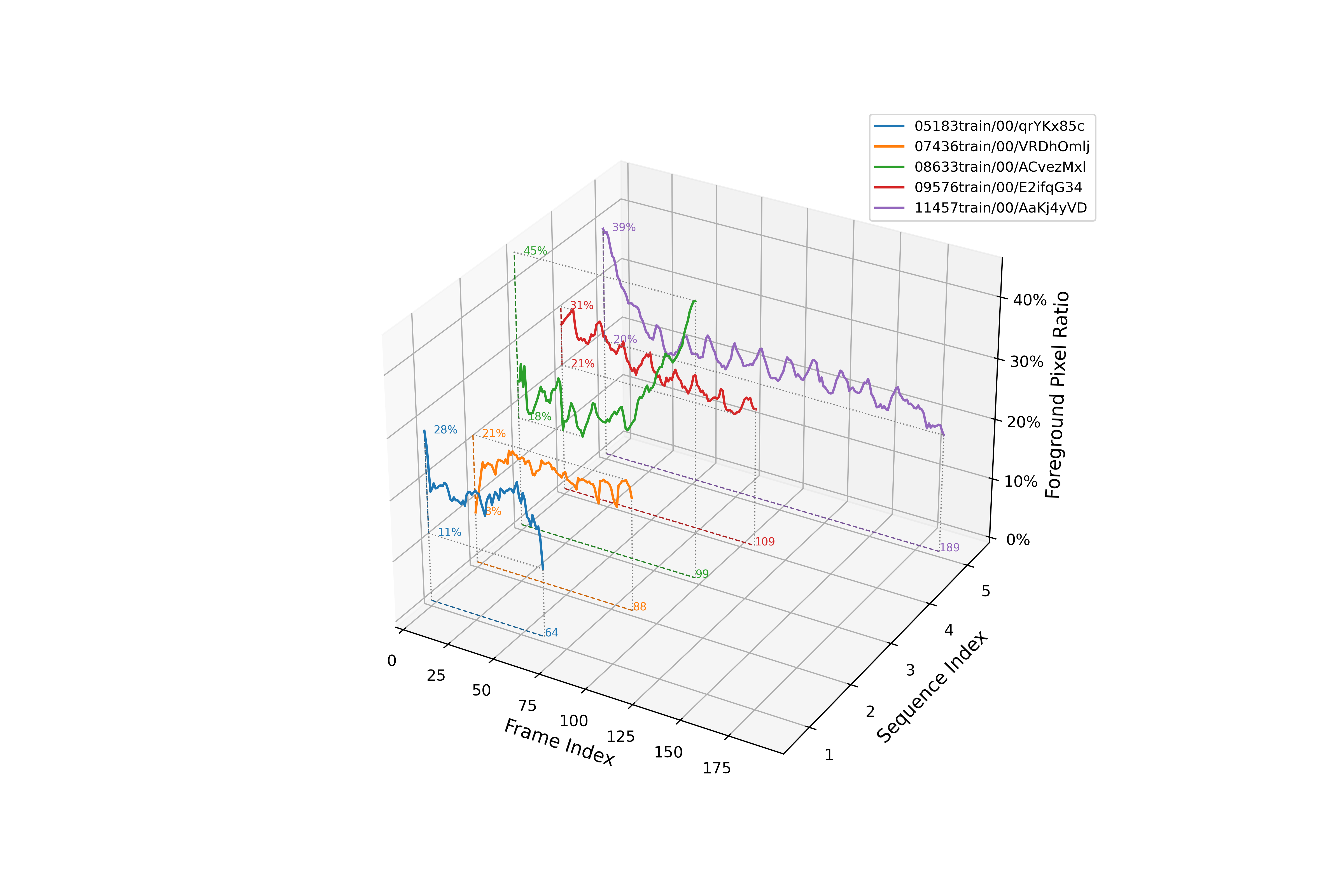}}
	\end{minipage}
	\hfill
	\begin{minipage}[b]{0.39\textwidth}
		\centering
		\begin{minipage}[b]{0.49\textwidth}
			\centering
			\subfloat[OU-MVLP]{\includegraphics[width=0.75\textwidth, trim=8cm 2.5cm 5.5cm 2.5cm, clip]{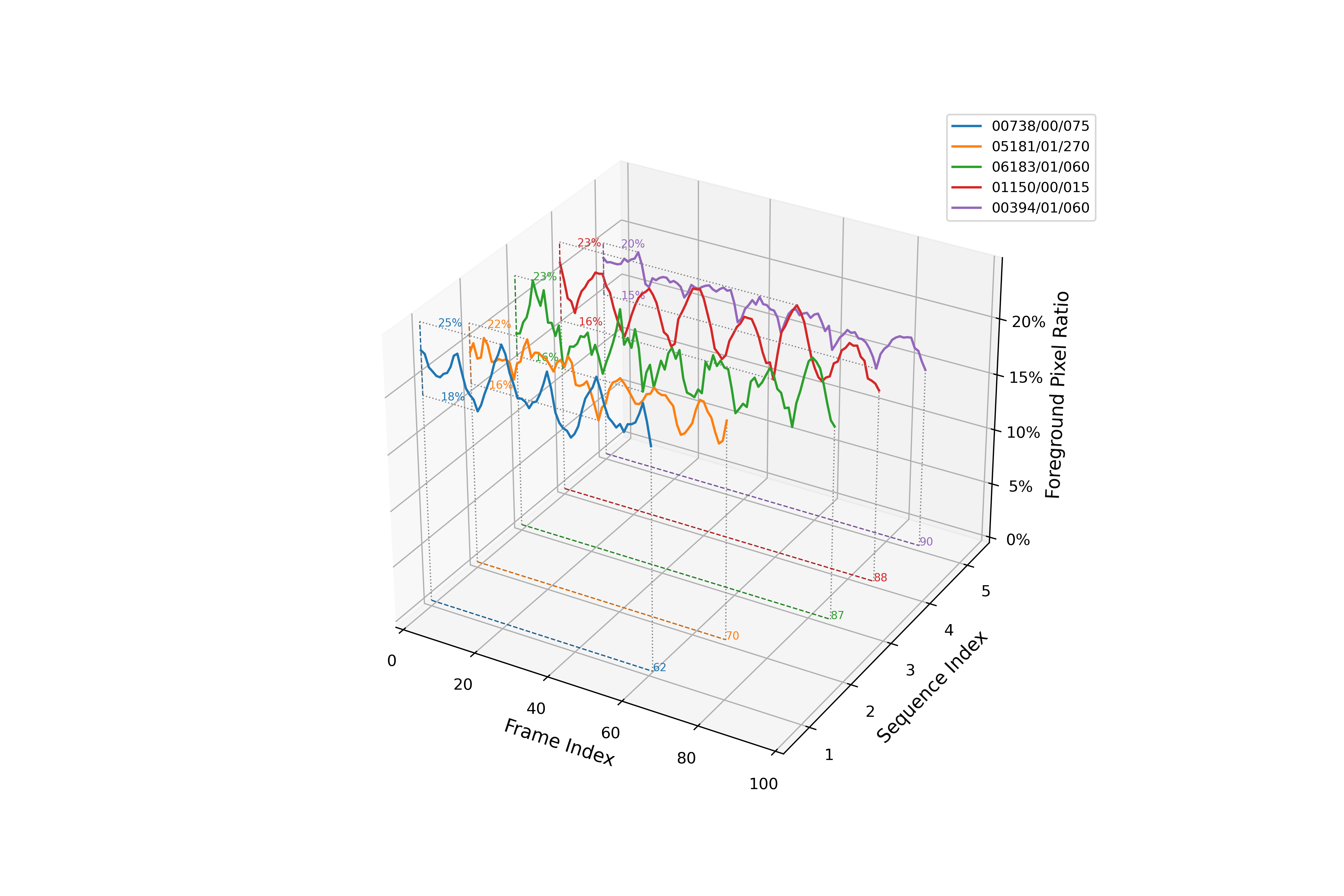}}
		\end{minipage}
		\hfill
		\begin{minipage}[b]{0.49\textwidth}
			\centering
			\subfloat[CASIA-B]{\includegraphics[width=0.75\textwidth, trim=8cm 2.5cm 5.5cm 2.5cm, clip]{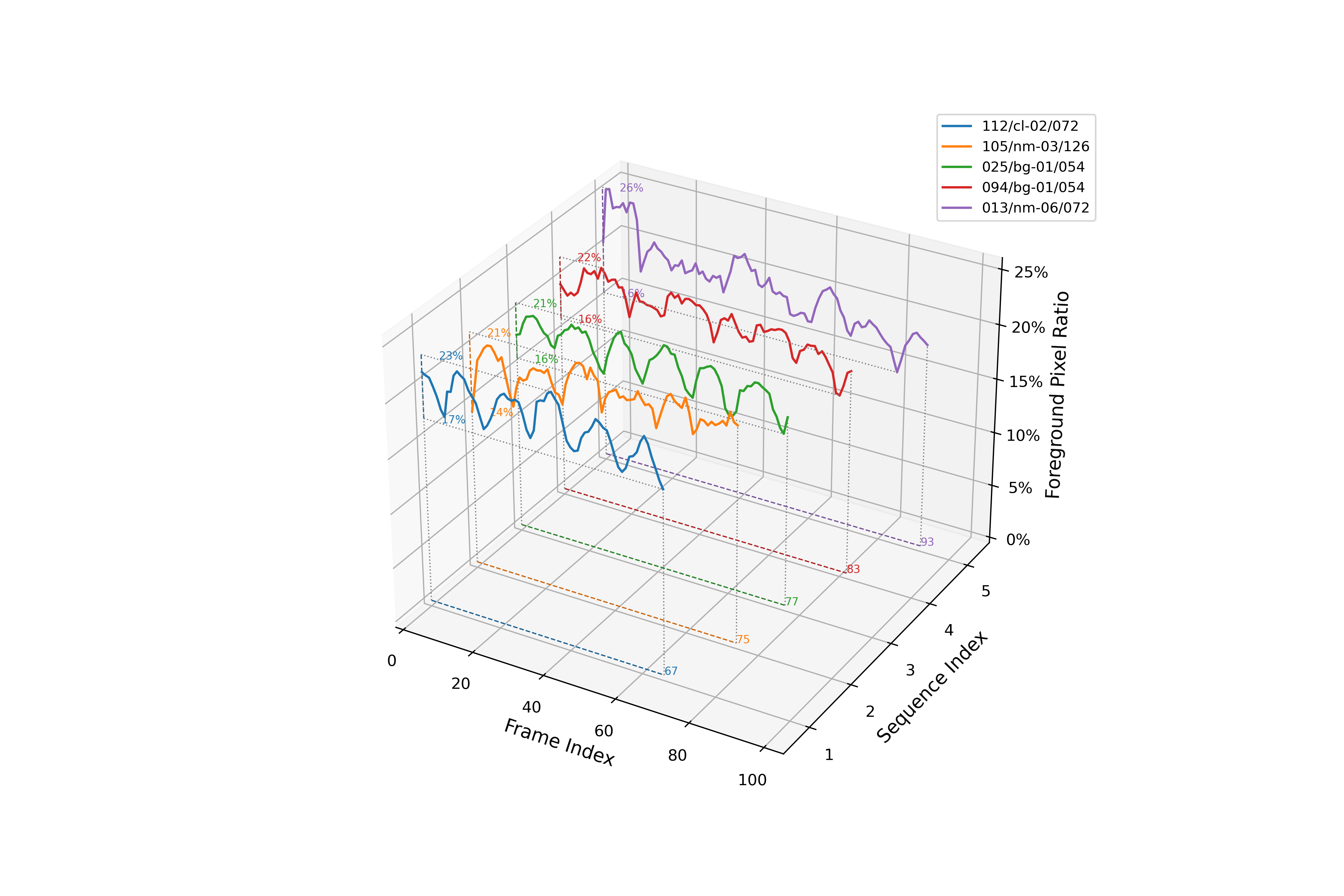}}
		\end{minipage}
		\begin{minipage}[b]{0.49\textwidth}
			\centering
			\subfloat[CCPG]{\includegraphics[width=0.75\textwidth, trim=8cm 2.5cm 5.5cm 2.5cm, clip]{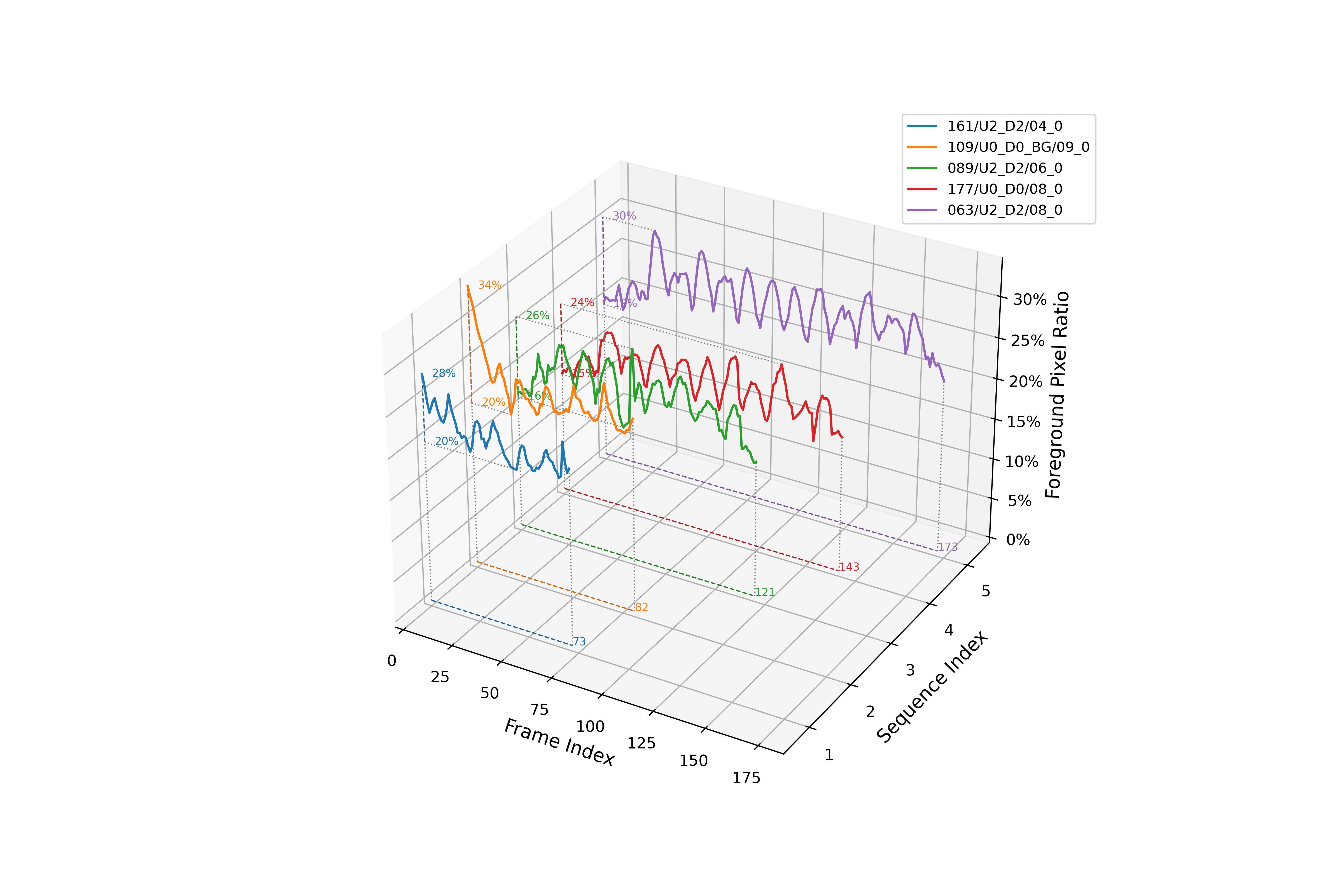}}
		\end{minipage}
		\hfill
		\begin{minipage}[b]{0.49\textwidth}
			\centering
			\subfloat[SUSTech1K]{\includegraphics[width=0.75\textwidth, trim=8cm 2.5cm 5.5cm 2.5cm, clip]{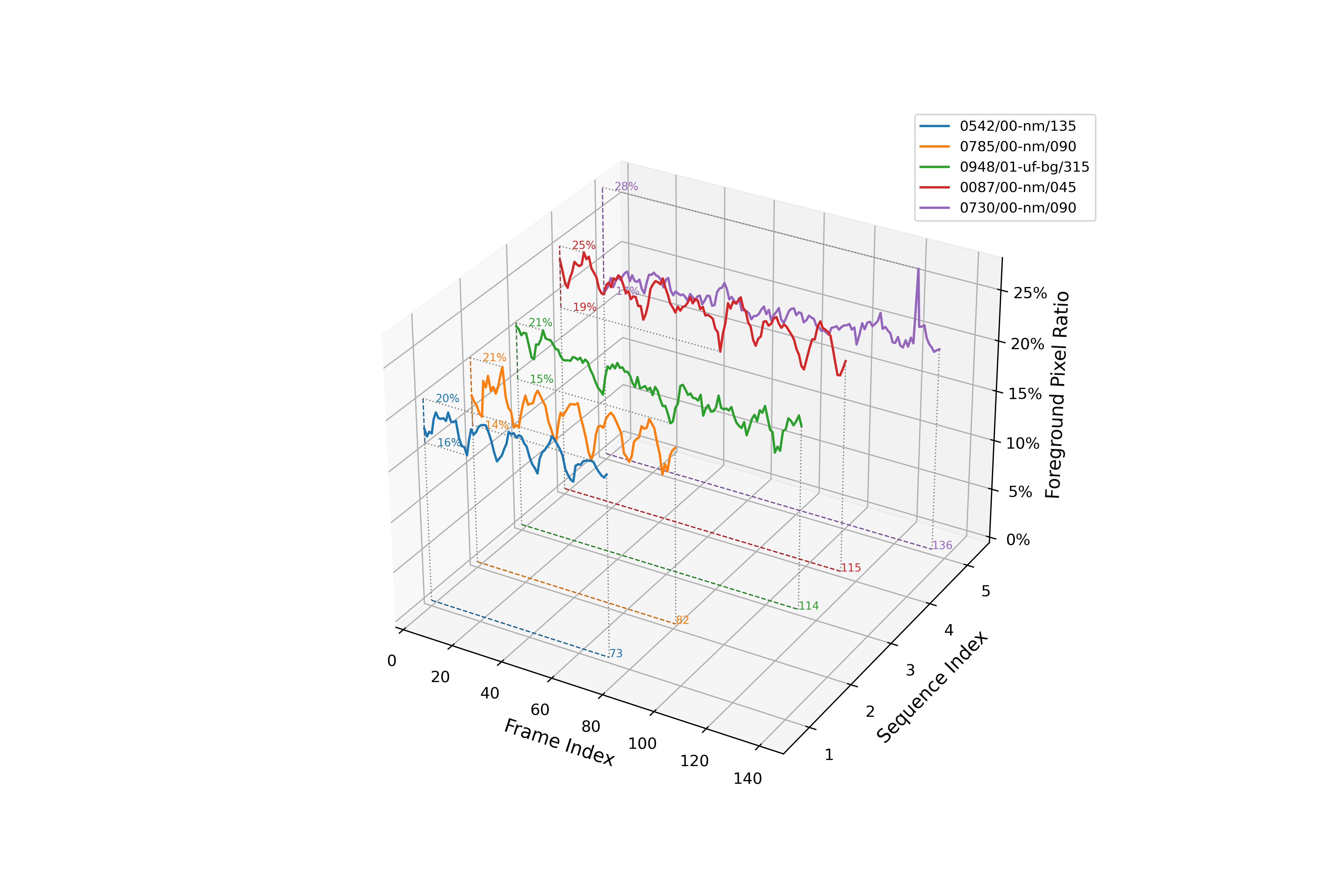}}
		\end{minipage}
	\end{minipage}
	\hfill
	
	\caption{{Foreground pixel rates on public gait datasets.}}
	\label{fig:pixelrate}
	%\vspace{-0.3cm}
\end{figure*}

To provide a more intuitive understanding of this observation, we randomly select five gait sequences from each dataset and plot the foreground pixel rates in Figure \ref{fig:pixelrate}. The periodicity of pedestrian silhouettes in the four constrained datasets is stable, whereas this is not the case for the wild datasets. Under such conditions, the estimation of gait cycles may not be accurate. To address this, we design a new metric, named AIFV, to calculate average inter-frame variation in a gait datasets. 

For each gait sequence, we calculate the foreground pixel rate $\text{p}_n$ for each silhouette:

\begin{equation}
	\text{p}_n = \frac{\sum_{x=1}^{h} \sum_{y=1}^{w} \mathbb{I}(I_n(x, y) > 0.5)}{h \times w},
\end{equation} 	
where $I_n$ denotes the pixel value at coordinates $(x, y)$ at frame $n$, $\mathbb{I}$ is an indicator function that returns 1 if the condition holds true indicating a foreground pixel, and 0 otherwise. $h$ and $w$ are the height and width of the silhouette, respectively.
Then, the maximum fluctuation of this percentage sequence $\{\text{p}_n\}$ is computed as a measure of variability for the sequence. Finally, on a dataset, AIFV score is calculated from the average of all variability measures from the training set, taken as a metric of the average inter-frame variation in the gait dataset:

\begin{equation}
	AIFV = \frac{1}{M} \sum_{i=1}^{M} (\max \{ \text{p}_n \}_{n=1}^{N} - \min \{ \text{p}_n \}_{n=1}^{N}),
\end{equation} 
where $M$ is the number of gait sequences in the training set of a dataset.

Lower AIFV scores (see Table \ref{tab:aifv}) indicate that the foreground pixel ratio fluctuates within a smaller range, suggesting that the sequence is more stable. As a result, gait cycle estimation is more accurate under these conditions. Since half gait cycles typically span 12-15 frames, a sampling length of 16 frames (RTS-$16$) appears to be optimal.

In contrast, when AIFV score increases, it suggests that the pedestrian walking scenarios are more complex. Therefore, it is essential to analyze the results based on the specific scene of the datasets. 
In Gait3D, pedestrians are indoors shopping, and their walking is frequently interrupted, leading to a lack of clear periodicity in the sequences. This can be seen as a reduction in the gait cycle from Figure \ref{fig:pixelrate}. Consequently, reducing the sampling length and increasing the number of random sampling positions may improve the ability to capture features across various walking states.
Meanwhile, in GREW, pedestrians walk outdoors. Since the camera is not strictly lateral, pedestrians often walk either toward or away from the camera, causing continuous changes in the gait images. This can be regarded as an extension of the gait cycle. In such cases, increasing the sampling length may enhance the extraction of discriminative features from the entire sequence.

\subsection{Impact of Training Iterations}

We train TrackletGait on Gait3D for 120,000 iterations, whereas DeepGaitV2-P3D-64 is trained for only 60,000 iterations. To ensure a fair comparison, we conducted experiments to examine the impact of training iterations (see Table \ref{tab:iteration}). DeepGaitV2-P3D-64 achieves a final performance of 74.4\% as reported in their paper~\cite{fan2023exploring} with 60,000 iterations, while we obtained a 74.6\% performance by extending the training to 120,000 iterations. The results suggest that additional training iterations can slightly improve performance on large datasets such as Gait3D. {Although the number of subjects in Gait3D is smaller than in GREW, the challenge in Gait3D persists. In Gait3D, the walking patterns of pedestrians are more complex compared to GREW, as pedestrians are engaged in shopping rather than simply walking. Consequently, increasing the number of training iterations may enhance the model's performance. Overall, under fair training conditions, TrackletGait still outperforms DeepGaitV2.}

\begin{table}[t]
	\setlength{\abovecaptionskip}{1pt}
	\renewcommand{\arraystretch}{1.3}
	\caption{Extensive experiments on {Gait3D} dataset considering training iterations.}
	\setlength{\tabcolsep}{2.8mm}
	\centering
	\begin{tabular}{c|c|c|c}
		\hline
		\renewcommand{\arraystretch}{1.2}\begin{tabular}[c]{@{}c@{}}\textbf{Exp.}\\ \textbf{Index}\end{tabular} &
		\textbf{Method} &
		\renewcommand{\arraystretch}{1.2}\begin{tabular}[c]{@{}c@{}}\textbf{Training}\\	\textbf{Iterations}\end{tabular}  & \renewcommand{\arraystretch}{1.2}\begin{tabular}[c]{@{}c@{}}\textbf{Rank-1}\\ \textbf{Accuracy (\%)}\end{tabular}  \\
		
		\hline
		\ding{172} & DeepGaitV2-P3D-64~\cite{fan2023exploring} & 60,000  & 74.4  \\
		\ding{173} & DeepGaitV2-P3D-64~\cite{fan2023exploring} & 120,000  & 74.6  \\
		\hline
		\ding{174} & TrackletGait-64 & 120,000  & \textbf{78.2}  \\
		\hline
	\end{tabular}
	\label{tab:iteration}
%	\vspace{-0.3cm}
\end{table}

\begin{table}[t]
	\setlength{\abovecaptionskip}{1pt}
	\renewcommand{\arraystretch}{1.3}
	\caption{Extensive experiments of P3D residual unit on {Gait3D} dataset. $n\times n$: 2d convolutional layer with $n\times n$ kernel size.}
	\setlength{\tabcolsep}{1.6mm}
	\centering
	\begin{tabular}{c|ccc|c}
		\hline
		\multirow{2}{*}{\renewcommand{\arraystretch}{1.2}\begin{tabular}[c]{@{}c@{}}\textbf{Exp.}\\ \textbf{Index}\end{tabular}} & \multicolumn{3}{c|}{\textbf{P3D Residual Unit}}  & \multirow{2}{*}{\renewcommand{\arraystretch}{1.2}\begin{tabular}[c]{@{}c@{}}\textbf{Rank-1}\\ \textbf{(\%)}\end{tabular}}             \\
		
		\cline{2-4} 
		
		& \textbf{Component-1} & \textbf{Component-2} & \textbf{Component-3} &  \\
		
		\hline
		\ding{172} & $3\times3$, stride 1  & Haar DWT & $1\times1$, stride 1    & \textbf{78.2}  \\
		\ding{173} & $3\times3$, stride 1  & $3\times3$, stride 2 & $1\times1$, stride 1   & 76.8  \\
		\ding{174} & $3\times3$, stride 1  & $3\times3$, stride 2 & -   & 77.2  \\
		\ding{175} & $3\times3$, stride 1  & MaxPool, stride 2 & - & 77.0 \\
		\ding{176} & $3\times3$, stride 2  & - & - & 75.8  \\
		\hline
		
	\end{tabular}
	\label{tab:spatialdownsampling}
	%\vspace{-0.3cm}
\end{table}

\subsection{Impact of Spatial Downsampling}
We conducted experiments on various downsampling methods, including strided convolution and max pooling, which are commonly used in previous work, as shown in Table \ref{tab:spatialdownsampling}. The position of each component is illustrated in Figure \ref{fig:unit}. In our residual unit, component-2 reduces spatial resolution while increasing the number of channels. Therefore, we introduce component-3, a $1\times1$ convolution layer, to restore the original number of channels. To ensure that the performance improvement is not due to this extra convolutional layer, we replaced the Haar DWT layer with a standard strided convolution while keeping component-3 unchanged. We also evaluate several different combinations, and the results demonstrate that the applied Haar DWT layer contributes to improved gait recognition in the wild.

\subsection{Impact of Temporal Branch}
\label{sec:TemporalBranch}

We conducted experiments to investigate the effects of kernel size and channel groups in 1D temporal convolution, as presented in Table \ref{tab:temporalbranch}. Based on the architecture of TrackletGait, we initially varied the kernel size while keeping the number of channel groups fixed at 4. As the kernel size increased from $3$ to $9$, the rank-1 accuracy exhibited a gradual improvement, peaking at 78.4\% with a $9$ kernel. However, a further increase in kernel size to $11$ resulted in a performance decline to 75.3\%. {We believe this is due to the RTS-$8$ sampling method we employed, where each tracklet has a length of 8. Therefore, using kernel sizes of 7 or 9 allows for sufficient tracklet information to be captured within a single temporal convolution. Based on this, we can draw a simple conclusion: larger convolutional kernel sizes help improve recognition performance. However, when the kernel size is increased to 11,} excessively large kernels may introduce excessive zero padding information, negatively affecting performance due to the training sequence length of 32 frames.

Next, we examine the effect of varying the number of channel groups while fixing the kernel size at $7$. A single-channel group achieves the highest rank-1 accuracy of 78.5\%, {as all channels contribute jointly to the convolutional computation, thereby improving the feature extraction capability. However, this also results in an increase in the number of parameters in the convolutional kernels}. As the number of groups increases, we observe a slight decline in accuracy with the reduction in parameters. Overall, our experiments demonstrate that kernel sizes and channel groups create a trade-off between model complexity and accuracy in gait recognition.

\begin{table}[t]
	\setlength{\abovecaptionskip}{1pt}
	\renewcommand{\arraystretch}{1.3}
	\caption{Extensive experiments of 1D temporal convolution on {Gait3D} dataset.}
	\setlength{\tabcolsep}{3.0mm}
	\centering
	\begin{tabular}{c|cc|c|c}
		\hline
		\multirow{3}{*}{\renewcommand{\arraystretch}{1.2}\begin{tabular}[c]{@{}c@{}}\textbf{Exp.}\\ \textbf{Index}\end{tabular}} & \multicolumn{2}{c|}{\textbf{Temporal Convolution}}  &

		\multirow{3}{*}{{\renewcommand{\arraystretch}{1.2}\begin{tabular}[c]{@{}c@{}}\textbf{\#Params}\\ \textbf{(M)}\end{tabular}}} &

		\multirow{3}{*}{\renewcommand{\arraystretch}{1.2}\begin{tabular}[c]{@{}c@{}}\textbf{Rank-1}\\ \textbf{Accuracy (\%)}\end{tabular}}             \\
		
		\cline{2-3} 
		
		& \renewcommand{\arraystretch}{1.2}\begin{tabular}[c]{@{}c@{}}\textbf{Kernel}\\ \textbf{Size}\end{tabular} & \renewcommand{\arraystretch}{1.2}\begin{tabular}[c]{@{}c@{}}\textbf{Channel}\\ \textbf{Group}\end{tabular} &  &  \\
		
		\hline
		\ding{172} & $3\times1\times1$   & 4 & \textbf{9.72}  & 77.7  \\
		\ding{173} & $5\times1\times1$   & 4 & 10.02 & 77.8  \\
		\ding{174} & $7\times1\times1$   & 4 & 10.31 & 78.2  \\
		\ding{175} & $9\times1\times1$   & 4 & 10.61 & \textbf{78.4}  \\
		\ding{176} & $11\times1\times1$ & 4 & 10.90 & 75.3  \\
		\hline
		\ding{177} & $7\times1\times1$   & 1 & 11.05 & \textbf{78.5}  \\
		\ding{178} & $7\times1\times1$   & 2 & 11.35 & 78.0  \\
		\ding{179} & $7\times1\times1$   & 4 & 10.31 & 78.2  \\
		\ding{180} & $7\times1\times1$   & 8 & \textbf{9.80} & 77.6 \\
		\hline
		
	\end{tabular}
	\label{tab:temporalbranch}
	%\vspace{-0.3cm}
\end{table}

\begin{table}[t]
	\setlength{\abovecaptionskip}{1pt}
	\renewcommand{\arraystretch}{1.3}
	\caption{{Extensive experiments of horizontal pooling on {Gait3D} dataset.}}
	\setlength{\tabcolsep}{2.9mm}
	\centering
	\begin{tabular}{c|c|c|c}
		\hline
		\textbf{Exp. Index} & \textbf{Module} & \textbf{Bin Number} & \textbf{Rank-1 Accuracy (\%)} \\
		\hline		
		\ding{172} & HP  & [16]         & \textbf{78.2} \\
		\ding{173} & HP  & [8]          & 78.1 \\
		\hline	
		\ding{174} & HPM  & [16, 8]      & 77.0 \\
		\ding{175} & HPM & [16,8,4,2,1] & 76.3  \\
		\hline
		
	\end{tabular}
	\label{tab:HorizontalPooling}
	\vspace{-0.3cm}
\end{table}

\begin{table}[t]
	\setlength{\abovecaptionskip}{1pt}
	\renewcommand{\arraystretch}{1.3}
	\caption{{Extensive experiments of loss functions on {Gait3D} dataset.}}
	\setlength{\tabcolsep}{2.2mm}
	\centering
	\begin{tabular}{c|c|c|c}
		\hline
		\textbf{Exp. Index} & \textbf{Triplet Loss} & $\boldsymbol{\alpha}$ & \textbf{Rank-1 Accuracy (\%)} \\
		\hline
		
		\ding{172} & HE Triplet   & 1/2   & 77.4 \\
		\ding{173} & HE Triplet   & 2/3   & \textbf{78.2}  \\
		\ding{174} & HE Triplet   & 3/4   & 77.6  \\
		\hline
		\ding{175} & Batch-all Triplet \cite{HermansBeyer2017Arxiv} & 1     & 77.2 \\
		\ding{176} & Batch-hard Triplet \cite{HermansBeyer2017Arxiv} & -     & 72.0 \\
		\hline
		
	\end{tabular}
	\label{tab:lossfunction}
	%\vspace{-0.3cm}
\end{table}

\subsection{{Impact of Horizontal Pooling}}

We conducted experiments to investigate the effects of Horizontal Pooling. This module was first proposed as Horizontal Pyramid Mapping (HPM)~\cite{Song.2019}, where a pyramid structure is employed to extract multi-level features. Subsequently, in GaitBase \cite{opengait}, this module was simplified to Horizontal Pooling (HP), extracting features at only one level to reduce computational cost. In this paper, we compare the differences between these approaches, as shown in Table~\ref{tab:HorizontalPooling}. The results indicate that HPM does not improve performance because our previous network is already capable of effectively extracting features at multiple scales. Therefore, employing a pyramid structure to combine features at different scales does not enhance their discriminative ability.

\subsection{{Impact of Loss Functions}}

We conducted experiments to investigate the effects HE Triplet Loss, as shown in Table \ref{tab:lossfunction}. In HE Triplet Loss, $\alpha$ serves as a distance threshold to determine which triplets contribute to the effective loss. When $\alpha=1$, the proposed HE Triplet Loss becomes equivalent to the Batch-all Triplet Loss. As $\alpha$ decreases, triplets with the largest anchor-to-positive distances are excluded from the loss calculation, effectively filtering out low-quality silhouette pairs. The best performance is achieved when $\alpha = 2/3$, resulting in a 1.0\% improvement over the Batch-all Triplet Loss. Finally, when applying Batch-hard Triplet Loss, which contrasts with HE Triplet Loss, the performance drops significantly from 77.2\% to 72.0\%. This decline suggests that hard triplets do not contribute effectively to gait recognition in real-world scenarios, as silhouettes are less discriminative due to noise.

In this study, we do not explicitly evaluate the quality of gait silhouettes. Instead, we directly exclude triplets where the anchor-positive distance is excessively large. To further demonstrate the impact of the HE Triplet Loss, we selected a model trained on Gait3D for 80,000 iterations and present several silhouette samples in Table \ref{fig:cases}. Additionally, we list the distances with $\alpha = 2/3$ as defined in Equation \ref{equ:alpha}, with the samples to be excluded highlighted in red. It is evident that low-quality silhouettes with large anchor-positive distances are excluded during the training phase, thereby improving overall performance. Although this approach may result in the removal of some potentially valuable hard triplets, which could reduce training efficiency, the experimental results demonstrate that our proposed method remains effective.

\begin{table*}[t]
	\footnotesize 
	\centering
	\setlength{\abovecaptionskip}{1pt}
	\renewcommand{\arraystretch}{1.1}
	\caption{{An illustrate of silhouettes and distances in a training batch on Gait3D.}}
	\setlength{\tabcolsep}{4.5mm}
	\centering
	\begin{tabular}{c|ccc}		
		\hline
		\textbf{Anchor} & \multicolumn{3}{c}{\textbf{Positive Samples}} \\
		\hline
		\includegraphics[width=0.20\textwidth]{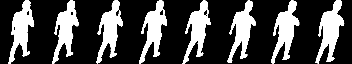} &
		\includegraphics[width=0.20\textwidth]{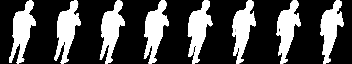} &
		\includegraphics[width=0.20\textwidth]{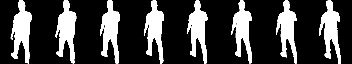} &
		\includegraphics[width=0.20\textwidth]{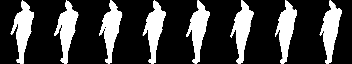} \\
		\textbf{ID: 0626} & $d_{ap} = 2.6860$ & $d_{ap} = 3.4202$ & $d_{ap} = 3.9598$ \\
		\hline
		\includegraphics[width=0.20\textwidth]{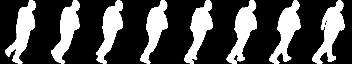} &
		\includegraphics[width=0.20\textwidth]{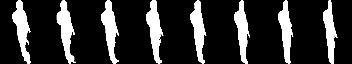} &
		\includegraphics[width=0.20\textwidth]{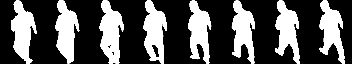} &
		\includegraphics[width=0.20\textwidth]{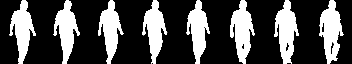} \\
		\textbf{ID: 2662} & \textcolor{red}{$d_{ap} = 4.4205>d$} & $d_{ap} = 3.6926$ & $d_{ap} = 3.7400$ \\
		\hline
		\includegraphics[width=0.20\textwidth]{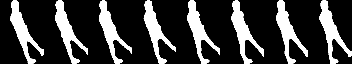} &
		\includegraphics[width=0.20\textwidth]{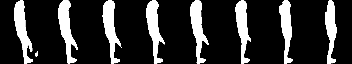} &
		\includegraphics[width=0.20\textwidth]{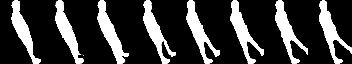} &
		\includegraphics[width=0.20\textwidth]{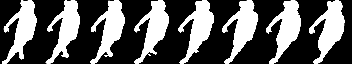} \\
		\textbf{ID: 2884} & \textcolor{red}{$d_{ap} = 4.3109>d$} & $d_{ap} = 2.5436$ & $d_{ap} = 3.8676$ \\
		\hline
		\includegraphics[width=0.20\textwidth]{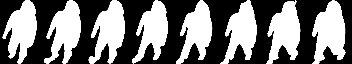} &
		\includegraphics[width=0.20\textwidth]{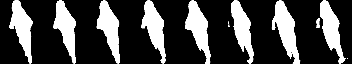} &
		\includegraphics[width=0.20\textwidth]{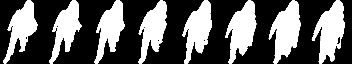} &
		\includegraphics[width=0.20\textwidth]{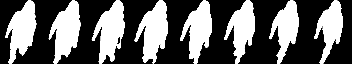} \\
		\textbf{ID: 2510} & $d_{ap} = 3.8335$ & $d_{ap} = 3.2294$ & $d_{ap} = 3.1602$ \\
		\hline
		\textbf{Distance}&\multicolumn{3}{c}{$d_{mean}=3.5720, d_{max}=4.4205, d=4.1377$} \\
		\hline
	\end{tabular}
	\label{fig:cases}
%	\vspace{-4mm}
\end{table*}

\subsection{{Discussion on Cross-dataset Results}}
\label{sec:Crossdataset}

Table \ref{tab:crossdatasets} presents the results of the cross-dataset experiments. In this setup, TrackletGait is trained on one dataset and directly evaluated on another, without fine-tuning or any other cross-domain adaptation methods. Due to the homogeneous acquisition scenarios of each gait dataset, the cross-dataset differences are substantial, resulting in a significant performance drop compared to single-dataset experiments. It is important to note that this work is not focused on cross-domain gait recognition; therefore, this experiment serves as a baseline for future cross-domain studies.

\begin{table}[t]
	\setlength{\abovecaptionskip}{1pt}
	\renewcommand{\arraystretch}{1.3}
	\caption{Performance on Cross-dataset Experiments.}
	\setlength{\tabcolsep}{0.65mm}
	\centering
	
	\begin{tabular}{c|c|c|c|c|c|c}
		\hline
		\multirow{2}{*}{\renewcommand{\arraystretch}{1.2}\begin{tabular}[c]{@{}c@{}}\textbf{Training}\\ \textbf{Dataset}\end{tabular}} & \multicolumn{6}{c}{\textbf{Testing Dataset, Rank-1 Accuracy (\%)}} \\ 
		\cline{2-7} 	
		& {\textbf{Gait3D}} & {\textbf{GREW}} & {\textbf{OU-MVLP}} & {\textbf{CASIA-B}} & {\textbf{CCPG}} & \textbf{SUSTech1K} \\ 
		\hline	
		\textbf{Gait3D}   &-&36.5&46.1&56.4&37.4&51.0\\
		\textbf{GREW}     &36.6&-&34.8&56.3&29.5&30.6\\
		\textbf{OU-MVLP}  &30.6&35.3&-&68.9&37.6&41.2\\
		\textbf{CASIA-B}  &17.3&21.4&30.8&-&33.6&43.0\\
		\textbf{CCPG}     &19.2&18.2&24.6&55.2&-&40.5\\
		\textbf{SUSTech1K}&21.1&18.3&30.3&65.9&33.4&-\\
		\hline 
	\end{tabular}
	
	\label{tab:crossdatasets}
	\vspace{-0.3cm}
\end{table}

%-------------------------------------------------------------------
\section{Conclusion}
\label{sec:conclude}

In this paper, we address the challenges of gait recognition in the wild. Existing methods do not account for the complex walking states present in real-world surveillance scenarios and fail to fully explore temporal sampling and spatial downsampling methods. To enhance the performance of gait recognition in the wild, we propose a novel framework named TrackletGait. TrackletGait employs Random Tracklet Sampling, a temporal sampling method that randomly samples short tracklets from a gait silhouette sequence, balancing robustness and representation. Additionally, we introduce Haar Wavelet-based Downsampling to preserve more discriminative information during spatial downsampling. Furthermore, we propose a Hardness Exclusion Triplet Loss to improve performance by excluding low-quality samples. Experimental results on the {six public gait datasets} demonstrate that TrackletGait outperforms previous state-of-the-art methods while maintaining a smaller network parameters. Consequently, TrackletGait provides a robust and efficient solution for gait recognition in the wild. In future work, we will continue to focus on real-world video surveillance to improve recognition performance and model interpretability \cite{chen2021explainable}.

\section*{Acknowledgments}
This work was supported by the National Nature Science Foundation of China (U21B2024), the Fundamental Research Funds for the Provincial Universities of Zhejiang (GK259909299001-044), the "Pioneer" and "Leading Goose" R\&D Program of Zhejiang Province(2024C01107, 2023C01030, 2023C03012), and the Research Funds of Hangzhou Dianzi University (KYS085624281).

\bibliographystyle{IEEEtran}
\bibliography{IEEEabrv,ref-trackletgait}

%\newpage


\begin{IEEEbiography}[{\includegraphics[width=1in,height=1.25in,clip,keepaspectratio]{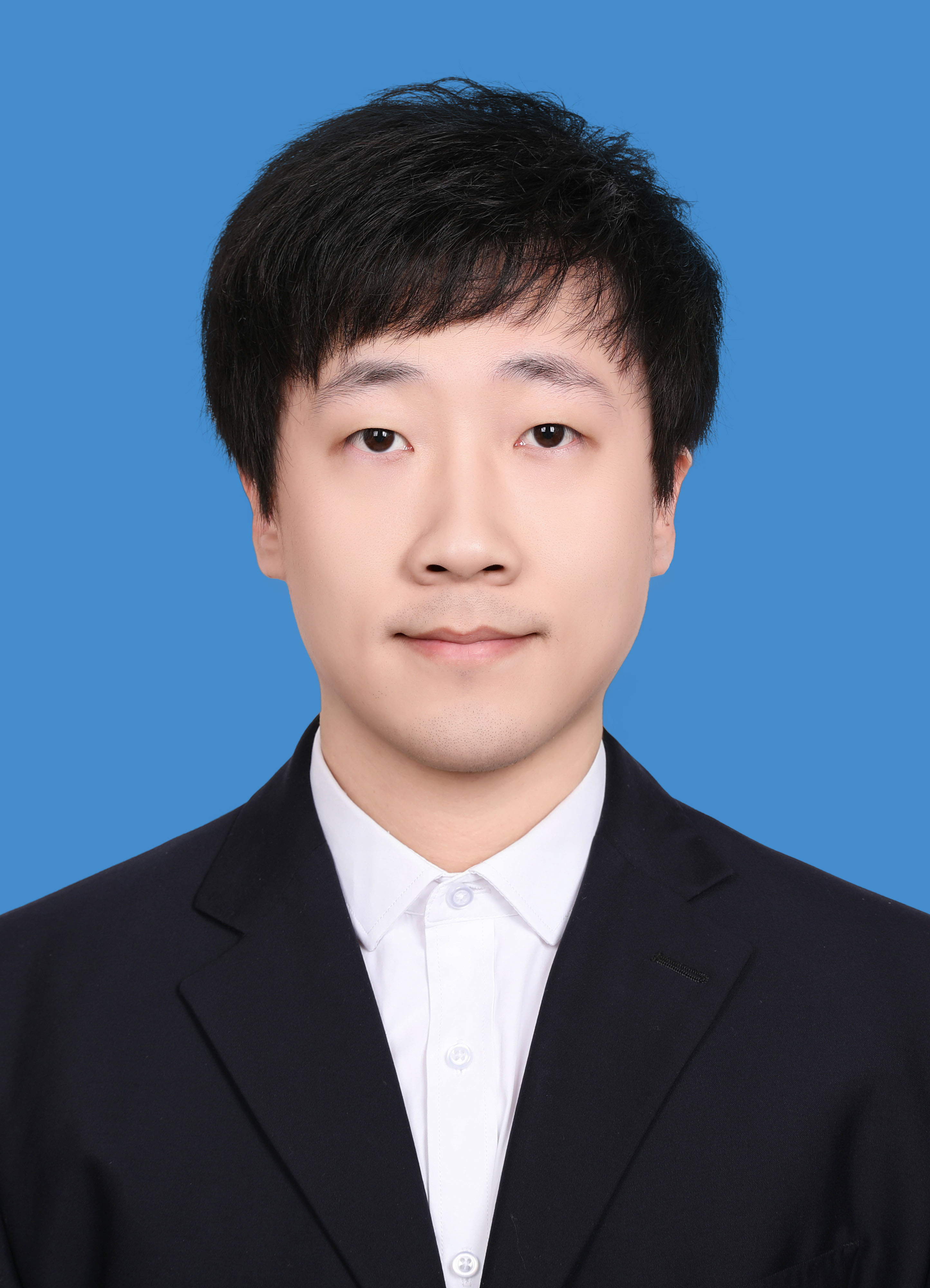}}]{Shaoxiong~Zhang} received the Ph.D. degree in Computer Science from Beihang University, Beijing, China, in 2023. He is currently an assistant professor at the School of Communication Engineering, Hangzhou Dianzi University, Hangzhou, China. His research interests include gait recognition, pattern recognition, and computer vision.
\end{IEEEbiography}

\begin{IEEEbiography}[{\includegraphics[width=1in,height=1.25in,clip,keepaspectratio]{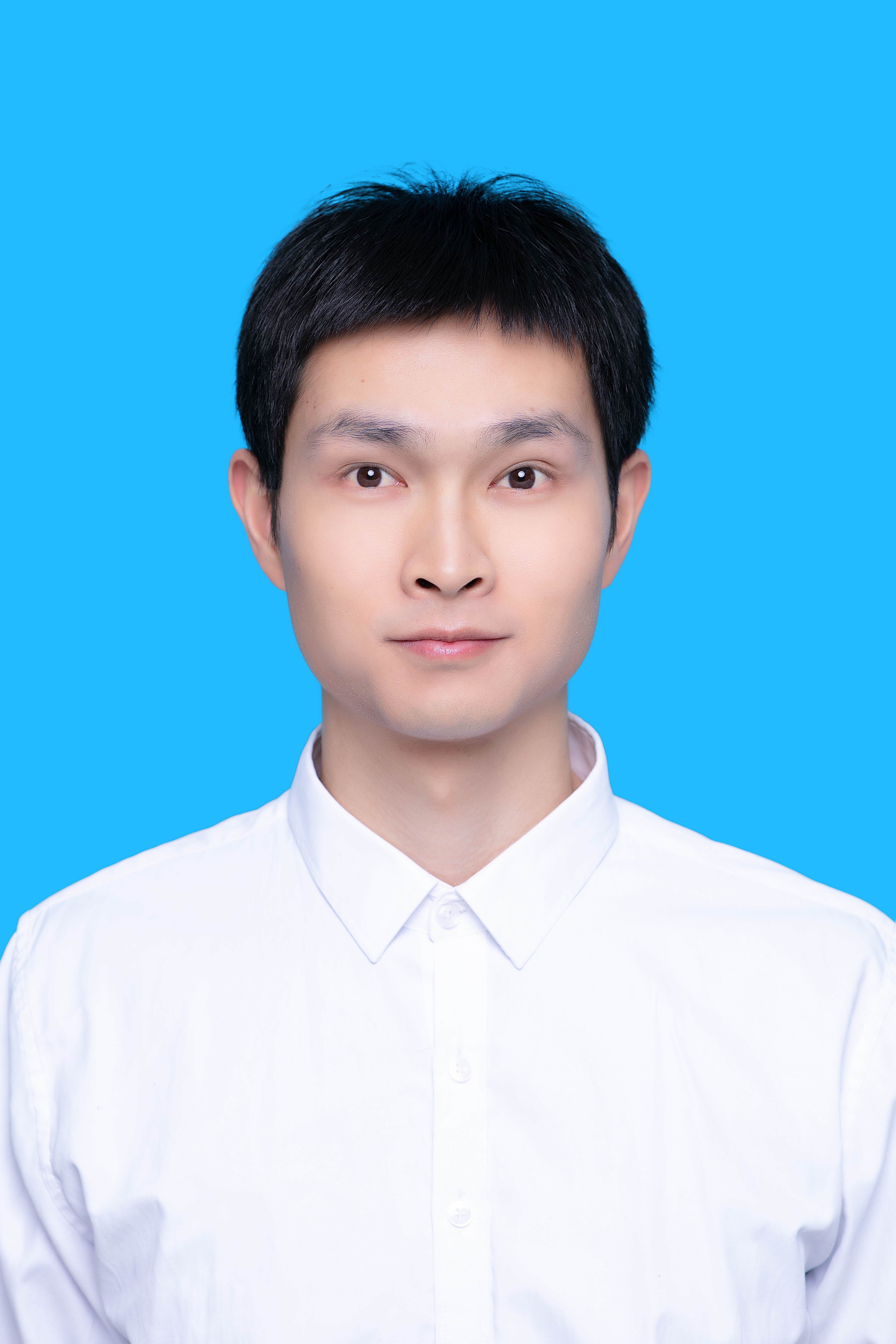}}]{Jinkai~Zheng} received the Ph.D. degree from Hangzhou Dianzi University in 2024. He is currently an associate professor with the School of Communication Engineering, Hangzhou Dianzi University, Hangzhou, China. He was a research algorithm engineer at JD Explore Academy, Beijing, China. His research interests include gait recognition and computer vision. His work on gait recognition has been published in IEEE CVPR, ACM MM, IEEE ISCAS, etc. He received the MSA-TC Best Paper Award - Honorable Mention at IEEE ISCAS in 2021.
\end{IEEEbiography}

\begin{IEEEbiography}[{\includegraphics[width=1in,height=1.25in,clip,keepaspectratio]{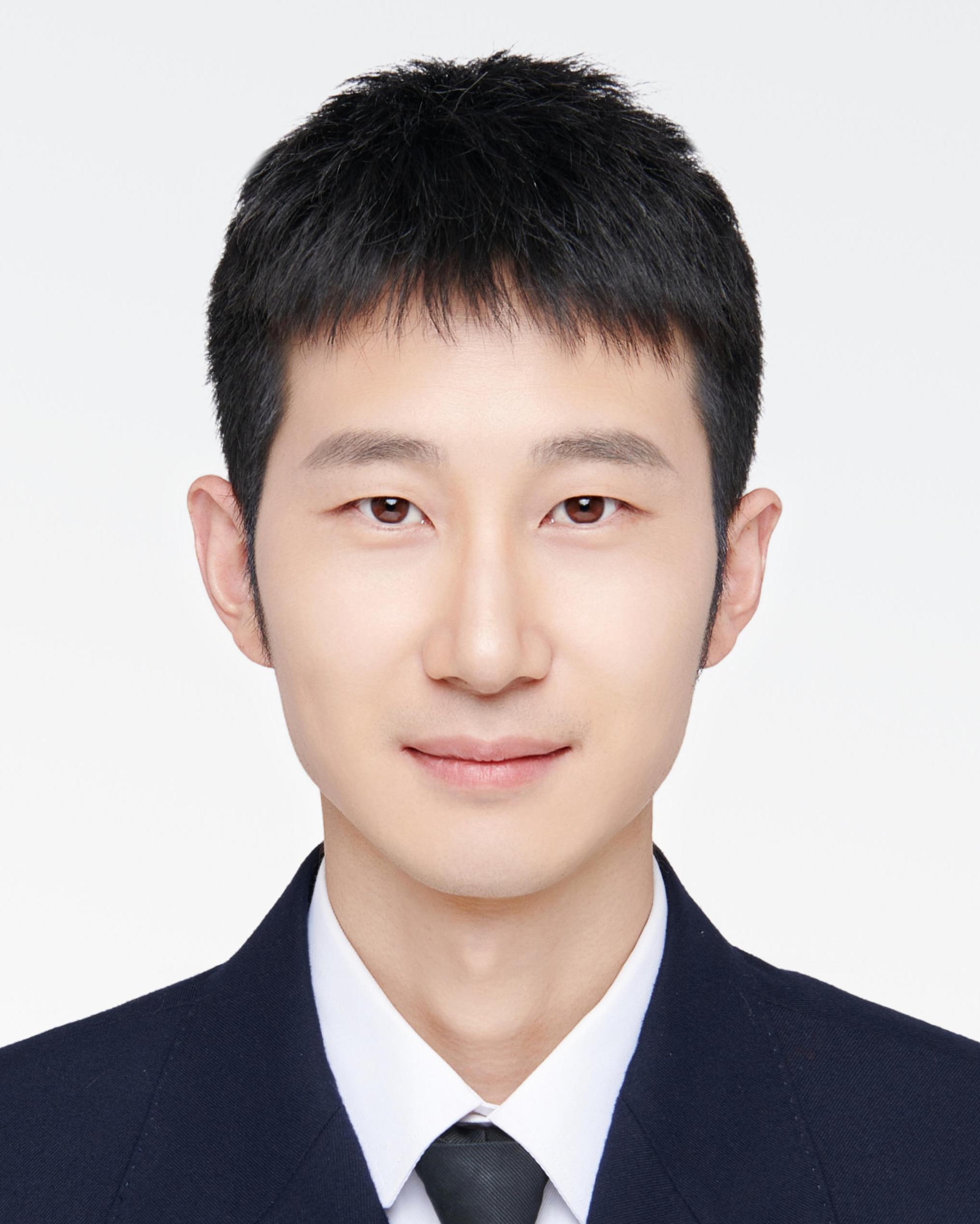}}]{Shangdong~Zhu} received the Ph.D. degree in robotics science and engineering from the Northeastern University, Shenyang, China, in 2024. He is currently a lecturer with the School of Communication Engineering, Hangzhou Dianzi University, Hangzhou, China. He is also a lecturer with the Intelligent Information Processing Laboratory, Hangzhou Dianzi University, Hangzhou, China. His research interests include computer vision, machine learning, and person re-identification.
\end{IEEEbiography}

\begin{IEEEbiography}[{\includegraphics[width=1in,height=1.25in,clip,keepaspectratio]{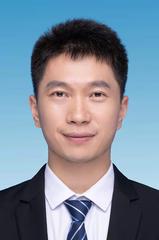}}]{Chenggang~Yan} received the B.S. degree in control science and engineering from Shandong University, Shandong, China, in 2008 and the Ph.D. degree in Computer Science from the Chinese Academy of Sciences University, Beijing, China, in 2013. He is a professor at the Hangzhou Dianzi University. His research interests include computational photography, pattern recognition, and intelligent systems.
\end{IEEEbiography}

\vfill

\end{document}